\documentclass[preprint,12pt]{elsarticle}

\usepackage{amssymb}
\usepackage{amsmath}
\usepackage{bm}
\usepackage{algorithmic}
\usepackage{algorithm}
\usepackage{multirow}
\usepackage{subfig}
\usepackage{url}
\usepackage{graphicx}
\usepackage{tikz}
\usetikzlibrary{shapes.geometric, arrows.meta, positioning, fit, calc}
\usepackage[colorlinks=true,linkcolor=blue,citecolor=blue,urlcolor=blue]{hyperref}

\journal{Preprint}

\begin{document}

\begin{frontmatter}

\title{ParkingTwin: Training-Free Streaming 3D Reconstruction for Parking-Lot Digital Twins}

\author[uestc]{Xinhao Liu}
\author[uestc]{Yu Wang}
\author[uestc]{Xiansheng Guo\corref{cor1}}
\ead{xsguo@uestc.edu.cn}
\author[xjlu]{Gordon Owusu Boateng}
\author[uestc]{Yu Cao}
\author[uestc]{Haonan Si}
\author[hebut]{Xingchen Guo}
\author[njit]{Nirwan Ansari}

\cortext[cor1]{Corresponding author.}

\affiliation[uestc]{organization={School of Information and Communication Engineering, University of Electronic Science and Technology of China},
             city={Chengdu},
             postcode={611731},
             country={China}}
\affiliation[xjlu]{organization={Department of Communications and Networking, Xi'an Jiaotong-Liverpool University},
             city={Suzhou},
             postcode={611731},
             country={China}}
\affiliation[hebut]{organization={School of Electrical Engineering, Hebei University of Technology},
             city={Tianjin},
             postcode={300401},
             country={China}}
\affiliation[njit]{organization={Advanced Networking Lab., Department of Electrical and Computer Engineering, New Jersey Institute of Technology},
             city={Newark},
             state={NJ},
             postcode={07102},
             country={USA}}

\begin{abstract}
High-fidelity digital twins of parking lots provide essential environmental priors for path planning, collision detection, and perception system validation of Automated Valet Parking (AVP). However, constructing such robot-oriented twins faces a fundamental "trilemma" involving geometric ambiguity, environmental interference, and computational constraints: 1) The restricted and sparse forward-facing views of mobile platforms lead to geometric degeneration in traditional  methods due to insufficient parallax; 2) Frequent dynamic occlusions (e.g., moving vehicles) and extreme lighting variations impede consistent texture fusion; and 3) Exis
ting neural rendering methods rely on computationally expensive offline optimization, failing to meet the real-time streaming requirements of edge-side robotics. To address these challenges, we propose ParkingTwin, a training-free, lightweight, and streaming 3D reconstruction system. The core innovations are three-fold: 1) OSM-Prior Driven Geometric Construction: We leverage OpenStreetMap (OSM) semantic topology to directly generate metric-consistent 3D Truncated Signed Distance Field (TSDF). This approache transforms "blind" geometric search into deterministic mapping, resolving the ill-posedness caused by sparse views while eliminating costly geometric optimization overhead. 2) Geometry-Aware Dynamic Filtering: We introduce a quad-modal geometric constraint field based on normal, height, and depth consistency to perform real-time rejection of dynamic vehicles and transient occlusions without prior training. 3) Illumination-Robust Fusion in the CIELAB Color Space: By incorporating adaptive L-channel weighting and depth gradient suppression, we decouple luminance and chromaticity in the perceptual space to eliminate seams and artifacts caused by abrupt lighting changes. Experiments demonstrate that our system achieves 30+ Frames Per Second (FPS) online streaming reconstruction on an entry-level GPU (GTX 1660). On a large-scale 68,000 $\text{m}^2$ real-world dataset, our method achieves an Structural Similarity Index Measure (SSIM) of 0.87 (a 16.0\% improvement), accelerates end-to-end processing by approximately 15$\times$, and reduces video memory usage by 83.3\% compared with state-of-the-art 3D Gaussian Splatting (3DGS) methods that require high-end GPUs (RTX 4090D). The system outputs explicit triangular meshes directly compatible with Unity/Unreal Engine (UE) digital twin workflows, effectively serving as an automated asset generator for initializing parking lot Digital Twins. Please visit our project page for the latest updates:
\url{https://mihoutao-liu.github.io/ParkingTwin/}.
\end{abstract}



\end{frontmatter}

\section{Introduction}

With the proliferation of Automated Valet Parking (AVP) and inspection robots, underground parking lots are becoming core application scenarios for parking lot digital twins~\cite{guo2025avpc,liu2025multi,boateng2025platform}, which must provide explicit geometry to support path planning and collision detection as well as photorealistic textures for perception system validation. Unlike traditional 3D visualization, digital twins in such scenarios impose stringent "dual requirements" on model quality: on the one hand, they must accurately represent explicit geometric structures such as column grids, parking lines, and driving lanes to support path planning and collision detection; on the other hand, they require high-fidelity textures reflecting real-world lighting distributions and material properties to facilitate closed-loop evaluation of autonomous driving perception algorithms. However, constructing such a system that balances geometric precision with visual fidelity, while supporting real-time edge-side updates for robots, presents severe technical challenges.

Existing scene modeling approaches generally fall into three technical paradigms, yet all struggle to directly meet the aforementioned demands. The first category, Neural Rendering methods (e.g., NeRF~\cite{mildenhall2020nerf}, Mip-NeRF~\cite{barron2021mip}, 3D Gaussian Splatting (3DGS)~\cite{kerbl20233d}), excels at generating high-quality novel views but mostly rely on implicit representations or unstructured point clouds. These methods suffer from difficulties in local editing, inability to directly remove transient occlusions, and prohibitive training costs. The second category, traditional Multi-View Stereo (MVS) reconstruction (e.g., OpenMVS~\cite{cernea2015openmvs}), can generate explicit meshes suitable for simulation platforms. However, in parking lots with uneven lighting and frequent dynamic occlusions, MVS is prone to texture seams and relies heavily on multi-view parallax. The third category, Neural Implicit Simultaneous Localization and Mapping (SLAM) (e.g., NICE-SLAM~\cite{zhu2022niceslam}, ESLAM~\cite{johari2023eslam}), supports online mapping and localization but remains dominated by high-compute "blind" geometric optimization. Their outputs are often implicit or semi-explicit, failing to meet the digital twin requirements for structural explicitness, editability, and real-time dynamic cleaning.

Crucially, parking lot scenarios are characterized by unique acquisition constraints, scene features, and environmental interferences, which collectively render existing methods ineffective. First, there is a dual constraint of acquisition mode and scene features: In practical inspections, robots are confined by traffic rules to perform unidirectional acquisition with sparse forward-facing views, resulting in severe parallax deficiency. Simultaneously, parking lots are dominated by large-scale textureless surfaces (e.g., floors, white walls) and repetitive geometric structures (e.g., identical columns). This adverse combination of "sparse views" and "feature paucity/ambiguity" causes traditional methods to fall into feature matching failures and geometric drift, leading to catastrophic structural collapse. Second, environmental inconsistency poses a major hurdle: Frequent occlusions by dynamic vehicles cause local information conflicts across views, which are easily "baked" into the model if not filtered. Furthermore, the interplay of artificial lighting and natural light from entrances creates extreme illumination gradients, making it difficult for traditional texture fusion algorithms to maintain consistency.

To address these challenges, we propose ParkingTwin, a training-free, streaming 3D reconstruction system tailored for parking lot digital twins. Our core insight is that learning geometry from scratch in man-made structured scenes is inefficient and unnecessary. We establish a new "prior-driven" paradigm that leverages the semantic topology from OpenStreetMap (OSM) to directly generate a metric-consistent 3D Truncated Signed Distance Field (TSDF) geometric skeleton. This strategy reduces the reconstruction task from "geometric estimation" to "texture mapping," fundamentally resolving the geometric ill-posedness caused by sparse views and weak textures, while eliminating expensive optimization overhead. Building on this function, we further propose a geometry-driven dynamic defense mechanism and an illumination-robust fusion algorithm in CIELAB color space~\cite{fairchild2013color}, achieving effective suppression of dynamic objects and lighting variations without training. The main contributions of this paper are summarized as follows:

\begin{enumerate}
\item \textbf{OSM Prior-Driven Real-Time Geometric Paradigm:} We introduce OSM semantics into the real-time reconstruction pipeline for the first time, utilizing semantic topology to directly generate 3D TSDF. This solves the problem of geometric collapse due to sparse views and weak textures while increasing geometric initialization speed by an order of magnitude.
\item \textbf{Perceptual Consistency Fusion in LAB Space:} We propose an adaptive L-channel weighting and depth gradient suppression strategy to decouple luminance and chromaticity in the perceptual color space, effectively eliminating seams and artifacts caused by extreme lighting gradients.
\item \textbf{Geometry-Driven Training-Free Dynamic Defense:} We construct a quad-modal geometric constraint field based on normal, height, edge, and depth consistency, enabling real-time rejection of dynamic vehicles and transient occlusions without deep learning training.
\item  \textbf{Efficient Streaming System and Open-Source Validation:} We propose a complete, lightweight streaming architecture capable of running on consumer-grade GPUs (e.g., GTX 1660). To foster reproducibility and future research in parking lot digital twinning, we fully open-source our codebase and the large-scale real-world dataset.
\end{enumerate}

\textbf{Organization:} Section 2 reviews related work (including an analysis of neural implicit SLAM methods); Section 3 details the methodology (covering OSM geometric initialization, training-free vehicle detection, LAB texture fusion, and system implementation); Section 4 presents experimental results (comparing with 3DGS and ESLAM, and demonstrating failure cases of traditional MVS methods like OpenMVS under sparse forward-facing acquisition); Section 5 discusses limitations and future directions.

\section{Related Work}

\subsection{3D Reconstruction and Texturing under Sparse Views}
Existing reconstruction methods face a dual failure of "parallax" and "computation" in parking lot scenarios. Traditional Multi-View Stereo (MVS) methods (OpenMVS~\cite{cernea2015openmvs}, COLMAP~\cite{schoenberger2016sfm}) rely on dense multi-view parallax for global graph-cut optimization. When inspection robots can only provide sparse forward-facing views, insufficient parallax renders depth matching an ill-posed problem, leading directly to extensive geometric collapse and holes. Geometric prior-driven reconstruction methods~\cite{HU2026124} have verified the critical role of prior knowledge in resolving geometric ill-posedness by transforming complex estimation problems into geometric matching tasks. Neural rendering methods (NeRF~\cite{mildenhall2020nerf}, Mip-NeRF~\cite{barron2021mip}, Instant-NGP~\cite{muller2022instant}, 3DGS~\cite{kerbl20233d}) achieve high-quality novel view synthesis via implicit or explicit Gaussian representations. Research in remote sensing also demonstrates the advantages of deep learning in large-scale reconstruction and change detection~\cite{zhu2017deep,nex2014uav,qin20163d}. However, the impact of occlusion on reconstruction quality cannot be overlooked. Petrovska \textit{et al.}~\cite{petrovska2025seeing} compared the reconstruction performance of MVS, NeRF, and Gaussian Splatting in vegetation-occluded scenarios, showing that MVS completeness drops sharply as occlusion increases, while NeRF maintains higher completeness but suffers from accuracy degradation proportional to the occlusion ratio.

The above-mentioned methods share three fundamental limitations: (1) They rely on expensive offline optimization, making them unsuitable for streaming workflows; (2) Their outputs are mostly implicit or unstructured point clouds, which are difficult to directly edit or export as standard meshes; (3) Geometry and appearance are entangled, making it difficult to distinguish dynamic objects from static backgrounds, thereby permanently "baking" vehicle ghosting into the texture. In structured scene reconstruction, hybrid-driven methods~\cite{WU2025100} have demonstrated the effectiveness of combining data-driven approaches with prior knowledge. Although targeted at specific regular structures, they validate the advantage of priors in such contexts. In contrast, our method leverages OSM semantic priors to directly generate explicit TSDF meshes, skipping geometric search and neural optimization to achieve real-time streaming processing while maintaining the editability of MVS.

\subsection{Neural Implicit SLAM and Semantic Priors} Neural Implicit SLAM (iMAP~\cite{sucar2021imap}, NICE-SLAM~\cite{zhu2022niceslam}, ESLAM~\cite{johari2023eslam}) attempts to introduce neural fields into real-time mapping. While classic dense/sparse SLAM (ORB-SLAM2~\cite{mur2017orb}, DROID-SLAM~\cite{teed2021droid}, Dense RGB-D SLAM~\cite{kerl2013dense}) is mature and efficient, it still relies on frame-by-frame feature or photometric optimization, making it difficult to maintain consistent geometry under sparse forward-facing views and dynamic occlusions. ESLAM reduces VRAM usage through multi-scale axis-aligned feature planes, yet it inherently remains a "blind geometric optimization" based on high-compute GPUs. It requires days to converge on a parking lot of tens of thousands of square meters, and its implicit geometry is difficult to use directly for collision detection or path planning.

Regarding semantic priors, OSM is primarily utilized for outdoor urban modeling~\cite{biljecki2017osm,goetz2012osm3d} or autonomous driving localization~\cite{schreiber2013lanelet} to generate coarse Level of Detail (LOD) models via extrusion rules, or as soft constraints to aid neural optimization~\cite{tancik2022block}. The importance of semantic information in complex scene understanding has been validated in various tasks; for instance, Light Detection and Ranging (LiDAR)-based place recognition~\cite{CHEN202497} enhances scene distinctiveness by fusing geometric and semantic relations, and indoor scene prior rule construction~\cite{jiang2023construction} demonstrates the effectiveness of building maps using scene-aware prior rules. In underground scene reconstruction, LiDAR point cloud-based end-to-end methods~\cite{LU2025563} show the efficacy of local feature aggregation, but such methods rely on deep learning training and mainly handle static point cloud data. Additionally, the application of mobile laser scanning data in urban environments~\cite{gehrung2022change} presents methods for handling localization errors and uncertainties, providing references for large-scale map updates. To the best of our knowledge, this work is the first to introduce OSM semantic topology into real-time geometric initialization for indoor/underground scenes, transforming the problem from ill-posed geometric estimation into deterministic mapping. This fundamentally overcomes both parallax and computational bottlenecks.

\subsection{Dynamic Scene Cleaning and Robust Texture Fusion} Dynamic object rejection typically relies on semantic segmentation (Mask R-CNN~\cite{he2017mask}, etc.) or motion consistency detection (DynaSLAM~\cite{bescos2018dynaslam}). The former requires massive annotated data and has an inference speed below 15 FPS, making it difficult to generalize to vehicles with diverse colors and shapes. The latter assumes continuous motion of dynamic objects, failing to handle transient "stop-and-go" occlusions in parking lots. In contrast, the geometric defense mechanism proposed in this paper is completely training-free, enabling robust rejection of dynamic vehicles at 30+ FPS using quad-modal constraints based on normal, height, edge, and depth consistency.

In terms of texture fusion, traditional methods~\cite{waechter2014let,zhou2018learning} perform weighted averaging in RGB space, where extreme lighting gradients cause luminance to dominate the weights, resulting in visible seams. Methods utilizing prior structural information to improve mesh quality in urban road reconstruction~\cite{zhu2021structure} also demonstrate the effectiveness of structure-aware completion. Building on this insight, we introduce luminance-chromaticity decoupling and depth gradient suppression in the LAB perceptual space, effectively eliminating lighting artifacts on non-Lambertian surfaces and accommodating the high dynamic range conditions typical of parking lot environments.

\section{Methodology}

\begin{figure*}[!t]
\centering
\includegraphics[width=\textwidth]{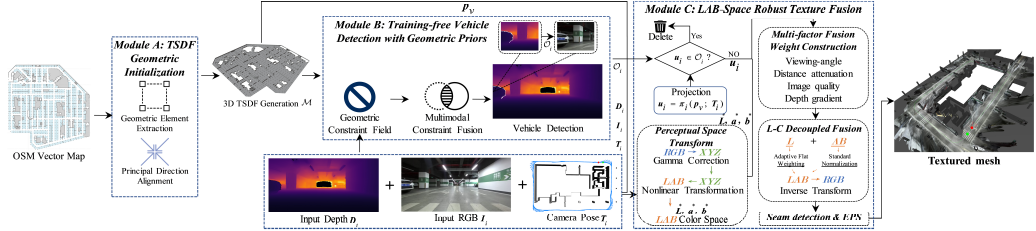}
\caption{The complete pipeline of the ParkingTwin system. The system operates in three stages: (1) \textbf{OSM-Prior Driven Geometric Initialization} directly generates a metric-consistent TSDF mesh from semantic priors; (2) \textbf{Geometry-Prior Based Dynamic Filtering} utilizes multi-modal constraints to remove vehicles without training; (3) \textbf{LAB Perceptual Fusion} ensures seamless texturing under varying illumination conditions.}
\label{fig:pipeline}
\end{figure*}

Figure~\ref{fig:pipeline} illustrates the comprehensive workflow of our system. Given an OSM map and an RGB-D sequence with camera poses, ParkingTwin is decomposed into three key modules: the Geometric Initialization Module $\mathcal{M}_\text{geo}$ (OSM-prior driven 3D TSDF initialization), the Dynamic Object Suppression Module $\mathcal{M}_\text{dyn}$ (geometry-prior based training-free vehicle detection and occlusion filtering), and the Illumination-Robust Texture Fusion Module $\mathcal{M}_\text{tex}$ (fusion in LAB perceptual space). Specifically, $\mathcal{M}_\text{geo}$ is responsible for constructing a topologically clean and metrically consistent explicit geometric framework; $\mathcal{M}_\text{dyn}$ utilizes multi-modal geometric cues upon this framework to identify and reject dynamic targets (e.g., vehicles) and untrustworthy observations; finally, $\mathcal{M}_\text{tex}$ performs illumination-robust texture fusion on the filtered set of valid observations to eliminate residual seams. Implementation details are elaborated in Sec.~\ref{sec:geo_module}, Sec.~\ref{sec:dyn_module}, and Sec.~\ref{sec:tex_module}, respectively.

\subsection{OSM-Prior Driven 3D TSDF Geometric Initialization}
\label{sec:geo_module}

\textbf{OSM-Driven Geometric Initialization.} To resolve the geometric ill-posedness inherent in sparse views, we forgo the noisy and computationally expensive traditional SfM/MVS pipeline. Instead, we leverage OSM semantic priors to directly generate a 3D TSDF. The OSM data used here is not coarse crowdsourced mapping, but high-precision topology extracted and vectorized from parking lot Computer-Aided Design (CAD) blueprints~\cite{feng2023osmag,cao2025osm2net}. To ensure metric consistency, we further employ LiDAR point clouds and vector maps for cross-source registration~\cite{qiu2025lidar}, correcting coordinate system deviations via feature consistency constraints. This ``CAD-level'' geometric prior provides the system with an absolutely precise scale and a noise-free topological skeleton, enabling instantaneous initialization of the geometric framework.

\subsubsection{OSM Geometry Extraction and Alignment}
We extract geometric elements (buildings, roads, parking areas, etc.) from the OSM map and align the OSM vector map to the metric system using known landmarks. To mitigate aliasing in subsequent TSDF generation, we extract the building edge directions $\theta_{\text{edge}}$, construct a histogram over $[0^\circ, 90^\circ)$, and rotate the map by $-\theta_{\text{dominant}}$ to align structures with the coordinate axes.

\subsubsection{3D TSDF Generation}
Parking structures exhibit vertical consistency. We calculate the Signed Distance Field $\Phi_{2D}(x, y)$ on the XY plane via Euclidean distance transform and then extrude the 2D profile along the Z-axis to a height $H$ to generate the 3D TSDF volume. The TSDF field function $\Phi(x,y,z)$ is truncated at $\pm\tau$, mapping the 3D space to signed distance values. To enhance the robustness of ground/roof extraction and noise suppression, we incorporate classic techniques such as cloth simulation filtering and individual tree segmentation during LiDAR preprocessing~\cite{zhang2016easy,li2012new}. The Marching Cubes algorithm is applied to this TSDF field to extract the zero-level set $\{\bm{p} \mid \Phi(\bm{p}) = 0\}$, yielding a triangular mesh $\mathcal{M} = (\mathcal{V}, \mathcal{F})$, where $\mathcal{V}$ is the set of vertices and $\mathcal{F}$ is the set of triangular faces. Subsequent texturing processes are performed on the vertices $\bm{p}_v \in \mathcal{V}$ of the mesh $\mathcal{M}$.

\subsection{Geometry-Prior Based Training-Free Vehicle Detection}
\label{sec:dyn_module}
\label{sec:vehicle_detection}

\subsubsection{Problem Formulation}
Given an RGB-D sequence $\{I_i, D_i\}$ and camera poses $\bm{T}_i$, our goal is to generate a binary mask $M_i$ for each frame to identify dynamic vehicles. We utilize the OSM-driven static mesh $\mathcal{M}$ as a geometric reference. By projecting $\mathcal{M}$ onto the current view $i$ via standard pinhole projection $\pi_i$, we obtain a reference static depth map $D_{\mathcal{M}}^i$, which serves as a baseline for inconsistency detection.

\subsubsection{Definition of Geometric Constraint Fields}
We define four complementary geometric constraint fields, each inducing a subset in the image domain $\Omega$ containing pixels that satisfy the respective constraint. Let $\pi_i^{-1}(\bm{u})$ denote the 3D point corresponding to pixel $\bm{u}$ (obtained via back-projection using depth $D_i(\bm{u})$), and let $\bm{n}_i(\bm{u})$ be the surface normal at that point (provided by the TSDF mesh $\mathcal{M}$).

\noindent\textbf{Normal Constraint Field.} Vehicles are parked on an approximately horizontal ground plane $\Pi_g: \bm{e}_z^T\bm{p} = z_g$ (where $\bm{e}_z = [0,0,1]^T$ is the vertical unit vector and $z_g$ is the ground height). The normal of their bottom contact surface should form an angle with $\bm{e}_z$ smaller than a threshold $\theta_g$. The normal constraint field is defined as:
\begin{equation}
\mathcal{C}_{\text{normal}}^i = \left\{ \bm{u} \in \Omega \;\Big|\; \bm{n}_i(\bm{u}) \cdot \bm{e}_z > \cos\theta_g \right\},
\label{eq:normal_constraint}
\end{equation}
where $\theta_g = 20^\circ$. This threshold is set according to parking ramp design standards (typically maximum slope $< 15^\circ$), with an additional margin of approximately $5^\circ$ to absorb normal estimation errors and sensor pose noise.

\noindent\textbf{Height Constraint Field.} Vehicles have a typical range of geometric dimensions. Their height relative to the ground, $h(\bm{p}) = \bm{p}_z - z_g$ (where $\bm{p}_z$ is the $z$-coordinate of the 3D point $\bm{p}$), should fall within $[h_{\min}, h_{\max}]$. The height constraint field is defined as:
\begin{equation}
\mathcal{C}_{\text{height}}^i = \left\{ \bm{u} \in \Omega \;\Big|\; h_{\min} \leq [\pi_i^{-1}(\bm{u})]_z - z_g \leq h_{\max} \right\},
\label{eq:height_constraint}
\end{equation}
where $[\pi_i^{-1}(\bm{u})]_z$ denotes the $z$-coordinate of the back-projected 3D point. $h_{\min} = 0.5$~m and $h_{\max} = 2.5$~m are derived from statistics of common passenger cars/SUVs: the 2.5~m upper limit covers most non-specialized vehicles while excluding ceiling fixtures, and the 0.5~m lower limit filters out low static appendages like curbs and speed bumps.

\noindent\textbf{Edge Discontinuity Constraint Field.} The depth field $D_i$ exhibits sharp transitions at the boundary between a vehicle and the ground, which can be characterized by the gradient norm. Let pixel coordinates be $\bm{u} = (u,v)$. We estimate the gradient $\nabla D_i(\bm{u}) \approx [D_i(u+1,v)-D_i(u-1,v), D_i(u,v+1)-D_i(u,v-1)]^T / 2$ using the Sobel operator. The edge constraint field is defined as:
\begin{equation}
\mathcal{C}_{\text{edge}}^i = \left\{ \bm{u} \in \Omega \;\Big|\; \|\nabla D_i(\bm{u})\| > \tau_{\text{edge}} \right\},
\label{eq:edge_constraint}
\end{equation}
where $\tau_{\text{edge}} = 1.0$~m/pixel. This constraint captures geometric discontinuities at vehicle boundaries.

\noindent\textbf{TSDF Depth Consistency Constraint Field.} The TSDF mesh $\mathcal{M}$ is generated from OSM priors and encodes only static scene geometry (excluding vehicles). Let $D_{\mathcal{M}}^i: \Omega \to \mathbb{R}^+$ be the theoretical depth map rendered from $\mathcal{M}$ using pose $\bm{T}_i$. When a vehicle is present, it acts as a foreground occlusion, causing the observed depth $D_i(\bm{u})$ to be strictly smaller than the theoretical depth $D_{\mathcal{M}}^i(\bm{u})$. The depth consistency constraint field is defined as:
\begin{equation}
\mathcal{C}_{\text{depth}}^i = \left\{ \bm{u} \in \Omega \;\Big|\; D_{\mathcal{M}}^i(\bm{u}) - D_i(\bm{u}) > \tau_{\text{depth}} \right\},
\label{eq:depth_constraint}
\end{equation}
where $\tau_{\text{depth}} = 0.3$~m. This value is derived from the system voxel resolution and typical extrinsic registration errors, serving as a robust tolerance (approximately $3\sigma$) to distinguish significant dynamic occlusions from sensor measurement noise.

\subsubsection{Multi-Modal Constraint Fusion}
The four constraint fields characterize the geometric properties of vehicles from complementary perspectives: normal and height constraints encode the spatial relationship of the vehicle relative to the ground; the edge constraint captures local geometric discontinuities; and the depth constraint leverages global prior references. We define the set of vehicle occlusion pixels as the intersection of the four constraint fields:
\begin{equation}
\mathcal{O}_i = \bigcap_{c \in \{\text{normal}, \text{height}, \text{edge}, \text{depth}\}} \mathcal{C}_c^i.
\label{eq:vehicle_detection}
\end{equation}

This intersection operation is equivalent to a pixel-wise joint decision:
\begin{equation}
m_i(\bm{u}) = \prod_{c \in \{\text{normal}, \text{height}, \text{edge}, \text{depth}\}} \mathbb{1}\left\{\bm{u} \in \mathcal{C}_c^i\right\},
\label{eq:indicator_fusion}
\end{equation}
where $\mathbb{1}\{\cdot\}$ denotes the indicator function that equals 1 when the condition is true and 0 otherwise.
This logical-AND fusion strategy requires all constraints to be satisfied simultaneously, significantly reducing the false positive rate (if a single constraint has a false positive rate $\alpha$, the joint rate drops to $\alpha^4$) while maintaining high recall. The detected vehicle occlusion pixels $\mathcal{O}_i$ are excluded during texture mapping, ensuring that only the static scene participates in fusion.

From a system perspective, the dynamic object suppression module $\mathcal{M}_\text{dyn}$ utilizes the aforementioned multi-modal geometric constraints to construct the vehicle occlusion set $\mathcal{O}_i$ for each frame, thereby defining the set of valid observations used in subsequent texture fusion (see Eq.~\eqref{eq:valid_indicator}). This module is responsible for both rejecting explicit vehicle pixels and filtering out unreliable occlusion observations via geometric/depth consistency. Its overall contribution is quantitatively and qualitatively evaluated in the cumulative ablation study in Section 4 via the comparison of Cfg.~A--D (see Sec.~\ref{sec:ablation_cumulative}).

\subsection{Illumination-Robust Texture Fusion in LAB Perceptual Space}
\label{sec:tex_module}

\subsubsection{Problem Statement and Motivation}
Parking lot scenarios exhibit extreme dynamic ranges: entrance areas are subjected to direct natural light ($>10^4$~lux), while deep windowless zones rely solely on sensor-activated artificial lighting ($\sim 10^2$~lux), resulting in exposure differences of 3--4 Exposure Value (EV) for the same surface across different viewpoints. Under such uncontrolled illumination, traditional RGB space fusion faces theoretical failure. Since RGB values couple material albedo with incident irradiance, direct weighted averaging ($\bm{c}_v = \sum w_i \bm{c}_i$) leads to \textit{Luminance Dominance}---where high-exposure frames numerically overwhelm low-exposure ones, forcing the fusion result to drift towards high-brightness observations. This systematic mathematical bias produces visible color banding and seam artifacts in lighting transition zones. To this end, we introduce the CIELAB perceptual space for texture fusion. Leveraging the physical decoupling of the luminance channel $L^*$ and chromaticity channels $a^*, b^*$ in LAB space~\cite{fairchild2013color}, we propose an adaptive $L^*$ weight modulation and depth gradient suppression strategy, thereby robustly handling extreme lighting gradients while preserving authentic material chromaticity.

\subsubsection{Perceptually Uniform Color Space Transformation}
To decouple luminance and chromaticity, we convert all RGB inputs to the CIELAB perceptually uniform color space~\cite{fairchild2013color}. The LAB space separates the luminance channel $L^*$ from the chromaticity channels $a^*, b^*$, such that the Euclidean distance in this space is proportional to human perceptual difference. This transformation allows us to treat the $L^*$ and $a^*, b^*$ channels separately, ensuring robust fusion under varying lighting conditions.

\subsubsection{Luminance-Chromaticity Decoupled Fusion Strategy}
Under the Lambertian assumption, the observed color of a surface in the $i$-th frame can be approximated as the product of material and illuminance:
\begin{equation}
\bm{c}_i = \rho \cdot E_i(\bm{n}, \bm{l}_i),
\label{eq:reflectance_model}
\end{equation}
where $\rho$ is the surface reflectance (an intrinsic material property) and $E_i(\bm{n}, \bm{l}_i)$ is the illuminance function (dependent on surface normal $\bm{n}$ and light source direction $\bm{l}_i$). Traditional weighted fusion in RGB space takes the form $\bm{c}_v^{\text{RGB}} = \sum_{i=1}^{N_v} w_i \bm{c}_i$ ($\sum w_i = 1$), where $N_v$ denotes the number of frames observing vertex $v$. When illuminance $E_i$ varies significantly, high-illuminance frames dominate the fusion result, masking the true material color and producing distinct seams in lighting transition zones.

A key property of the LAB space is that the $L^*$ channel primarily reflects scene illumination, whereas the $a^*, b^*$ channels largely capture surface intrinsic albedo~\cite{fairchild2013color}. Illuminance differences manifest mainly as variations in the $L^*$ component, while the $a^*, b^*$ components remain relatively insensitive to illuminance perturbations. Based on this observation, we adopt differentiated fusion strategies for the $L^*$ channel and the chromaticity channels.

Let a vertex $\bm{p}_v$ on the TSDF mesh $\mathcal{M}$ be mapped to the pixel $\bm{u}_i = \pi_i(\bm{p}_v)$ in the $i$-th frame via the projection function $\pi_i$. To exclude dynamic vehicle occlusions, we define a validity indicator function:
\begin{equation}
\mathbb{1}_{\text{valid}}^i(\bm{p}_v) = \begin{cases}
1, & \text{if}\ \pi_i(\bm{p}_v) \notin \mathcal{O}_i, \\
0, & \text{if}\ \pi_i(\bm{p}_v) \in \mathcal{O}_i,
\end{cases}
\label{eq:valid_indicator}
\end{equation}
where $\mathcal{O}_i \subset \Omega$ is the vehicle occlusion mask for the $i$-th frame (obtained from the multi-modal geometric detection in Sec.~\ref{sec:vehicle_detection}). The set of valid frames observing vertex $\bm{p}_v$ is denoted as $\mathcal{I}_v = \{i \mid \mathbb{1}_{\text{valid}}^i(\bm{p}_v) = 1\}$, with cardinality $N_v = |\mathcal{I}_v|$.

For each frame in the valid set $\mathcal{I}_v$, we construct a multi-factor fusion weight $w_i(\bm{p}_v) \in [0,1]$, incorporating factors such as viewing incidence angle, observation distance, image quality, and depth gradient (see details in Sec.~\ref{sec:view_weight}). While traditional methods apply normalized weights uniformly to all RGB channels, we adopt a differentiated weight normalization strategy tailored to the physical interpretation of the LAB space: chromaticity channels $a^*, b^*$ use standard normalized weights to fully leverage high-quality views for preserving true material color; the luminance channel $L^*$ employs power-law flattened weights to mitigate the excessive dominance of high-illuminance frames. The specific definitions and fusion process are detailed in the following subsections.

\subsubsection{Quality-Gated Multi-Factor Weight Construction}
\label{sec:view_weight}

To handle sparse and heterogeneous input observations, we construct a comprehensive weight $w_i(\bm{p})$ for each valid observation by integrating four quality factors. Let the unit normal vector of surface point $\bm{p} \in \mathcal{M}$ be $\bm{n}(\bm{p})$, the camera center of the $i$-th frame be $\bm{o}_i$, the normalized view direction be $\bm{v}_i(\bm{p}) = (\bm{o}_i - \bm{p})/\|\bm{o}_i - \bm{p}\|$, the incidence angle be $\theta_i(\bm{p}) = \arccos(\bm{n}(\bm{p}) \cdot \bm{v}_i(\bm{p}))$, and the observation distance be $d_i(\bm{p}) = \|\bm{o}_i - \bm{p}\|$.

\noindent\textbf{View Angle Weight $w_{\theta}(\theta_i)$.} As $\theta_i \to \pi/2$ (grazing observation), the projected pixel density tends to zero, causing severe distortion and blurring. To mitigate this effect, we employ a Hann window function
\begin{equation}
w_{\theta}(\theta) = \begin{cases}
1, & \theta < 30^\circ, \\
\frac{1}{2}\left[1 + \cos\left(\pi \frac{\theta - 30^\circ}{45^\circ}\right)\right], & 30^\circ \leq \theta \leq 75^\circ, \\
0, & \theta > 75^\circ,
\end{cases}
\label{eq:angle_weight}
\end{equation}
where $\theta_{\max} = 75^\circ$ is the upper bound for acceptable viewing angles.

\noindent\textbf{Distance Decay Weight $w_{d}(d_i)$.} We use a Lorentzian decay function
\begin{equation}
w_{d}(d) = \frac{1}{1 + (d/d_0)^2},
\label{eq:dist_weight}
\end{equation}
where $d_0 = 5$~m is the reference distance (corresponding to the decay point where $w_{d} = 0.5$).

\noindent\textbf{Image Quality Weight $w_{q}(q_i)$.} Let the normalized image quality score for the $i$-th frame be $q_i \in [0, 1]$ (derived from Laplacian variance and histogram statistics~\cite{cernea2015openmvs}). We define
\begin{equation}
w_{q}(q) = q^{\alpha_q}, \quad \alpha_q = 1.5,
\label{eq:quality_weight}
\end{equation}
where the exponent $\alpha_q > 1$ amplifies quality differences, further suppressing the contribution of low-quality frames.

\noindent\textbf{Depth Gradient Weight $w_{g}(g_i)$.} At occlusion boundaries, the depth field exhibits sharp transitions, leading to color misalignment. Let the depth gradient norm be $g(\bm{u}) = \|\nabla D_i(\bm{u})\|$ (estimated via Sobel operator). We define
\begin{equation}
w_{g}(g) = \begin{cases}
1, & g < 0.5~\text{m/pixel}, \\
\exp\left(-2.0 (g - 0.5)^2\right), & g \geq 0.5~\text{m/pixel},
\end{cases}
\label{eq:depth_weight}
\end{equation}
to effectively filter out contaminated colors at occlusion boundaries.

\noindent\textbf{Multi-Factor Weight Synthesis.} The comprehensive weight is constructed in a multiplicative form
\begin{equation}
w_i(\bm{p}) = w_{\theta}(\theta_i(\bm{p})) \cdot w_{d}(d_i(\bm{p})) \cdot w_{q}(q_i) \cdot w_{g}(g_i(\pi_i(\bm{p}))).
\label{eq:combined_weight}
\end{equation}
This multiplicative form implements a quality gating mechanism: if any single factor is zero, the overall weight becomes zero, thereby enforcing hard constraint exclusion. All factors can be computed independently and then multiplied, with a computational complexity of $\mathcal{O}(N_v)$.

Note that $w_i(\bm{p})$ given by Eq.~\eqref{eq:combined_weight} is an unnormalized weight, where only relative magnitudes reflect quality ranking. To convert it into convex combination coefficients (probability measure), we perform standard normalization over the valid frame set $\mathcal{I}_v$ (defined in Eq.~\eqref{eq:valid_indicator}):
\begin{equation}
\tilde{w}_i(\bm{p}) = \frac{w_i(\bm{p})}{\sum_{j \in \mathcal{I}_v} w_j(\bm{p})}, \quad \sum_{i \in \mathcal{I}_v} \tilde{w}_i(\bm{p}) = 1,
\label{eq:normalized_weight}
\end{equation}
where $\bm{p}$ refers to a specific mesh vertex $\bm{p}_v$. This standard normalized weight $\tilde{w}_i(\bm{p}_v)$ constitutes the convex combination coefficient for the fusion of chromaticity channels $a^*, b^*$ in Eq.~\eqref{eq:lab_blend}.

For the $L^*$ luminance channel, to avoid excessive dominance by high-illuminance frames, we employ a power-law flattening strategy. The modulation weight for the luminance channel is defined as
\begin{equation}
\tilde{w}_{L,i}(\bm{p}) = \frac{w_i(\bm{p})^{\gamma_L}}{\sum_{j \in \mathcal{I}_v} w_j(\bm{p})^{\gamma_L}}, \quad 0 < \gamma_L < 1,
\label{eq:l_weight}
\end{equation}
where $\gamma_L$ is the luminance weight flattening exponent (set to $\gamma_L = 0.5$ in this work). The power function $w_i(\bm{p})^{\gamma_L}$ maps weights to a flatter distribution (e.g., $w_i = 0.9 \to w_i^{0.5} = 0.95$, $w_i = 0.1 \to w_i^{0.5} = 0.32$), compressing the gap between high and low weights and balancing the contribution of each frame to luminance. This flattening strategy effectively suppresses the undue influence of overexposed or high-intensity frames on the final texture luminance.

\subsubsection{Weighted Fusion and Color Reconstruction}
Based on the normalized weight $\tilde{w}_i(\bm{p}_v)$ and the luminance flattening weight $\tilde{w}_{L,i}(\bm{p}_v)$, we perform luminance-chromaticity decoupled fusion in the CIELAB color space over the valid frame set $\mathcal{I}_v$. The fused color for vertex $\bm{p}_v$ in LAB space is calculated using only the valid frame set $\mathcal{I}_v$:
\begin{equation}
\bm{c}_v^{\text{LAB}} = \begin{bmatrix} L_v^* \\ a_v^* \\ b_v^* \end{bmatrix} = \begin{bmatrix} \sum_{i \in \mathcal{I}_v} \tilde{w}_{L,i}(\bm{p}_v) L_i^* \\ \sum_{i \in \mathcal{I}_v} \tilde{w}_i(\bm{p}_v) a_i^* \\ \sum_{i \in \mathcal{I}_v} \tilde{w}_i(\bm{p}_v) b_i^* \end{bmatrix},
\label{eq:lab_blend}
\end{equation}
where $\bm{c}_i^{\text{LAB}} = [L_i^*, a_i^*, b_i^*]^T$ is the LAB color of the $i$-th frame at that vertex. This fusion strategy ensures that dynamic vehicle pixels are completely excluded via the validity indicator $\mathbb{1}_{\text{valid}}^i$ (Eq.~\ref{eq:valid_indicator}), so the result reflects only the intrinsic material properties of the static scene. The fused result is then transformed back to RGB space via the inverse LAB-to-RGB transform for final texture mapping:
\begin{equation}
\bm{c}_v^{\text{RGB}} = \mathcal{T}_{\text{LAB}\to\text{RGB}}^{-1}(\bm{c}_v^{\text{LAB}}).
\label{eq:lab_to_rgb}
\end{equation}

This completes the full color processing pipeline from RGB input to LAB fusion and back to RGB output, achieving illumination-robust multi-view texture fusion.

\subsubsection{Seam Detection and Edge-Preserving Smoothing}
Although the aforementioned LAB perceptual fusion strategy (Eq.~\eqref{eq:lab_blend}) significantly improves texture consistency through luminance-chromaticity decoupling and multi-factor weighting, mesh discretization can result in adjacent triangular faces having different sets of valid visible frames $\mathcal{I}_v$. Variations in the optimal viewpoint combination can introduce color discontinuities at face boundaries, forming algorithmic seam artifacts. Unlike intrinsic color transitions at real material boundaries, these seams are algorithmic artifacts that require post-processing removal. However, simple global smoothing blurs real material edges, reducing texture fidelity. To remove these seams without blurring real material boundaries, we propose a variance-based seam detection followed by edge-preserving smoothing strategy.

\noindent\textbf{Variance-Based Seam Detection Criterion.} Let $\bm{v}_j \in \mathcal{V}$ ($j = 1, \ldots, |\mathcal{V}|$) be a mesh vertex, and let its set of adjacent faces be $\mathcal{N}(\bm{v}_j) = \{f_k \in \mathcal{F} \mid \bm{v}_j \in f_k\}$, where $\mathcal{F}$ is the set of triangular faces. We define the local color variance statistic at vertex $\bm{v}_j$ as
\begin{equation}
\sigma_j^2 = \frac{1}{|\mathcal{N}(\bm{v}_j)|} \sum_{f_k \in \mathcal{N}(\bm{v}_j)} \|\bm{c}_{f_k} - \bar{\bm{c}}_j\|_2^2,
\label{eq:color_variance}
\end{equation}
where $\bm{c}_{f_k} \in \mathbb{R}^3$ is the RGB color of face $f_k$ (converted from the LAB fusion result via Eq.~\eqref{eq:lab_to_rgb}), $\bar{\bm{c}}_j = \frac{1}{|\mathcal{N}(\bm{v}_j)|}\sum_{f_k \in \mathcal{N}(\bm{v}_j)} \bm{c}_{f_k}$ is the average color of the neighborhood, and $\|\cdot\|_2$ denotes the Euclidean norm. This variance statistic $\sigma_j^2$ quantifies color dispersion within the vertex neighborhood: real material edges typically form continuous high-variance curves in space (corresponding to physical material boundaries), whereas algorithmic seams manifest as isolated high-variance vertices (due to local inconsistencies from different viewpoints). Based on this observation, the set of seam vertices is defined as
\begin{equation}
\mathcal{V}_{\text{seam}} = \left\{\bm{v}_j \in \mathcal{V} \;\Big|\; \sigma_j^2 > \tau_{\text{seam}}\right\},
\label{eq:seam_vertices}
\end{equation}
where the threshold $\tau_{\text{seam}} = 100$ (corresponding to a relative color difference of approximately 3.9\% for RGB range $[0, 255]$). This threshold is determined via cross-validation on a validation set to balance false positive and false negative rates.

\noindent\textbf{Edge-Preserving Bilateral Smoothing.} For the detected set of seam vertices $\mathcal{V}_{\text{seam}}$, applying isotropic smoothing (e.g., Gaussian filtering) would blur real material boundaries. To simultaneously achieve the dual goals of seam elimination and edge preservation, we employ bilateral filtering for adaptive smoothing~\cite{tomasi1998bilateral}. The core idea of bilateral filtering is to consider both spatial proximity and range similarity: neighbors that are spatially close but have large color differences are assigned low weights, thereby preventing improper smoothing across real edges. Specifically, the smoothed color $\bm{c}_j'$ of a seam vertex $\bm{v}_j \in \mathcal{V}_{\text{seam}}$ is calculated as:
\begin{equation}
\bm{c}_j' = \frac{1}{Z_j} \sum_{f_k \in \mathcal{N}(\bm{v}_j)} w_{s}(d_{jk}) \cdot w_{r}(\Delta c_{jk}) \cdot \bm{c}_{f_k},
\label{eq:bilateral_filter}
\end{equation}
where $w_{s}$ is the spatial kernel and $w_{r}$ is the range kernel. The spatial kernel is defined as a Gaussian function based on the geometric distance $d_{jk} = \|\bm{v}_j - \bm{p}_{f_k}\|_2$ ($\bm{p}_{f_k}$ is the barycenter of face $f_k$):
\begin{equation}
w_{s}(d) = \exp\left(-\frac{d^2}{2\sigma_s^2}\right),
\label{eq:spatial_kernel}
\end{equation}
where $\sigma_s > 0$ is the spatial scale parameter controlling the neighborhood influence range. The range kernel is based on color difference $\Delta c_{jk} = \|\bm{c}_j - \bm{c}_{f_k}\|_2$:
\begin{equation}
w_{r}(\Delta c) = \exp\left(-\frac{(\Delta c)^2}{2\sigma_c^2}\right),
\label{eq:range_kernel}
\end{equation}
where $\sigma_c > 0$ is the range scale parameter controlling color tolerance. The normalization factor $Z_j = \sum_{f_k \in \mathcal{N}(\bm{v}_j)} w_{s}(d_{jk}) \cdot w_{r}(\Delta c_{jk})$ ensures Eq.~\eqref{eq:bilateral_filter} is a convex combination. The edge-preserving property of this bilateral filter is manifested as follows: when the color difference $\Delta c_{jk}$ between a neighbor face $f_k$ and the central vertex $\bm{v}_j$ is significantly larger than $\sigma_c$ (i.e., $\Delta c_{jk} \gg \sigma_c$), the range weight $w_{r}(\Delta c_{jk}) \approx 0$, suppressing the contribution of that face to smoothing and thus protecting real material boundaries; conversely, when $\Delta c_{jk} \ll \sigma_c$, $w_{r}(\Delta c_{jk}) \approx 1$, allowing the neighbor face to fully participate in smoothing, effectively eliminating seams. Regarding parameter selection, spatial scale $\sigma_s = 0.1$~m corresponds to 2--3 times the average mesh edge length, ensuring sufficient neighborhood coverage; range scale $\sigma_c = 15$ (RGB range $[0, 255]$) corresponds to approximately 6\% relative color difference, slightly higher than the seam detection threshold to avoid being overly conservative.

\subsection{Experimental Setup}

\textbf{Dataset and Acquisition Platform.} We conducted real-time reconstruction experiments in a real-world underground parking environment using a mobile robot platform and have fully open-sourced the resulting \textbf{ICPARK} dataset to facilitate reproducibility. The parking lot features a multi-level underground structure, with a single floor area of approximately $255~\mathrm{m} \times 267~\mathrm{m}$ ($\sim$68,000 $\text{m}^2$) and containing 1,397 standard parking spaces. RGB-D data was acquired using a ZED Stereo Camera (resolution $1280 \times 720$). Robot poses were derived by fusing linear velocity from wheel odometry with angular velocity and linear acceleration from an Inertial Measurement Unit (IMU) via temporal integration. \textbf{It is crucial to emphasize that the data acquisition strictly adhered to constraints typical of real-world inspection and autonomous driving scenarios:} the robot moved unidirectionally along driving lanes at a normal speed ($\sim$5 km/h), generating trajectories with \textit{sparse forward-facing views}, rather than the dense multi-angle continuous capture typical in traditional 3D reconstruction. While this acquisition mode better reflects practical deployment requirements, it results in surfaces being observed by only a few frames from similar viewpoints, imposing significantly higher demands on the reconstruction algorithm. The scene's OSM map provided building contours and parking space semantics for geometric initialization and scale unification (see Sec.~\ref{sec:geo_module} for the OSM acquisition process). The fully open-sourced dataset includes stereo image sequences, pose trajectories, OSM maps, and annotated vehicle masks to foster research on parking lot digital twins.

\begin{figure}[!t]
\centering
\includegraphics[width=0.6\columnwidth]{\detokenize{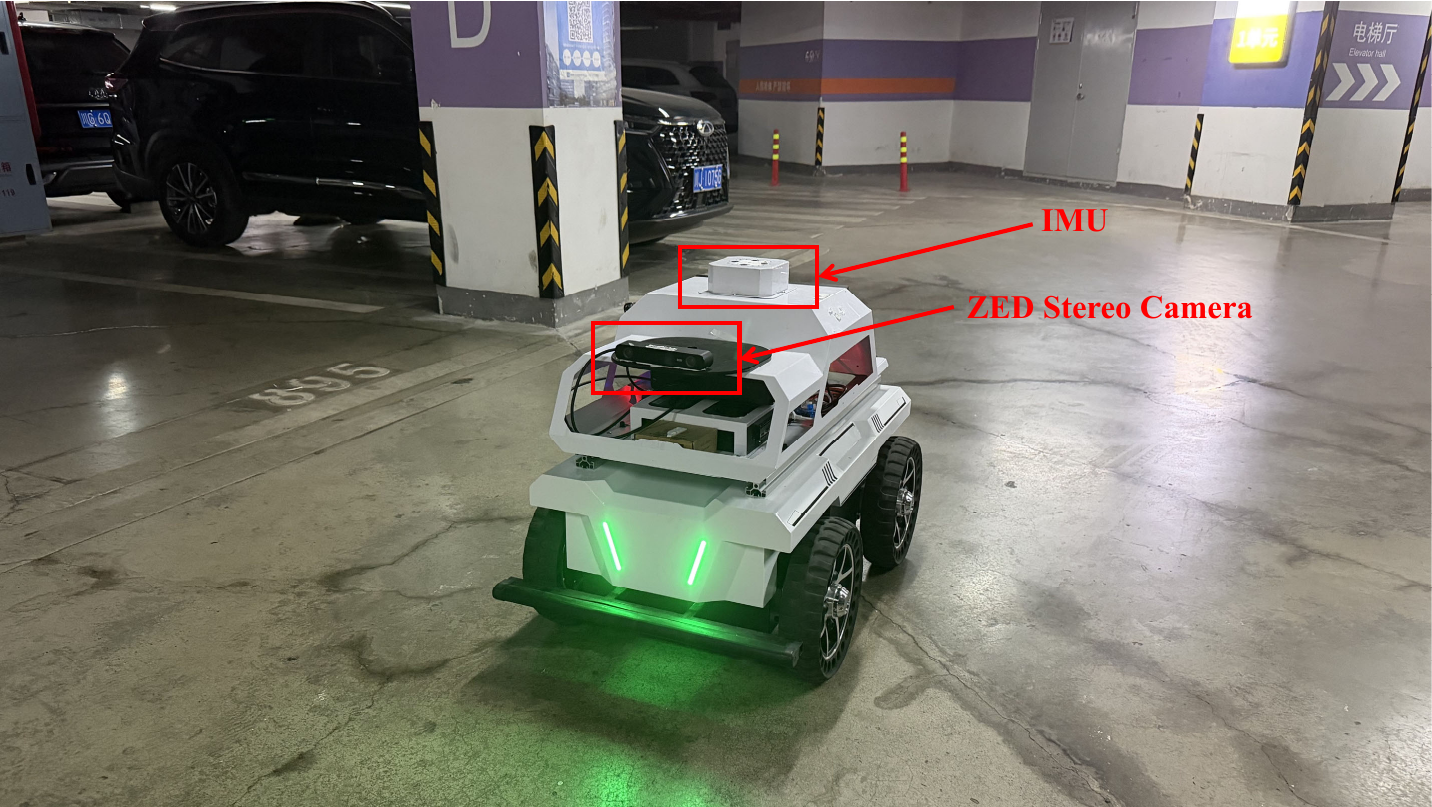}}
\caption{Dynamic challenges in the ICPARK dataset. Captured frames exhibit \textbf{high occlusion rates} (>40\% screen space) and diverse vehicle appearances. These factors render traditional semantic segmentation slow and unstable, necessitating our geometry-based removal strategy.}
\label{fig:vehicle_samples}
\end{figure}

\textbf{Implementation Platform and Operation Modes.} The system was implemented on a hardware platform equipped with an NVIDIA GTX 1660 (6GB VRAM), an Intel i9-12900K (16 cores), and 32GB DDR4 RAM. The software environment utilized Python 3.9, PyTorch 1.13, CUDA 11.7, Open3D 0.16 (for point cloud and mesh processing), and OpenCV 4.7 (for image processing and color space conversion). The system supports two operation modes: (1) \textit{Offline Batch Mode:} Performs post-processing reconstruction on the complete acquisition sequence to generate the final high-quality model; (2) \textit{Online Streaming Mode:} Based on the pre-generated OSM-driven TSDF geometric framework, the robot executes real-time vehicle detection and texture fusion during navigation. This ``reconstruct-while-moving'' capability supports incremental updates and real-time preview, meeting the needs of dynamic scene monitoring and rapid mapping. In online mode, the system maintains a processing frame rate of 30+ FPS.

\textbf{Baseline Methods.} We selected 3DGS and ESLAM as representatives of the blind reconstruction paradigm. All methods received identical RGB-D sequences as input and directly utilized the pose data. This comparison aims to demonstrate the cross-paradigm performance disparity: under the ill-posed constraints caused by sparse forward-facing views, baseline methods struggle to deconstruct geometry relying solely on visual cues, whereas ParkingTwin successfully breaks this data limitation by incorporating OSM priors.
\begin{itemize}
        \item \textbf{OpenMVS}~\cite{cernea2015openmvs}: A traditional multi-view texture mapping method.
        \item \textbf{3D Gaussian Splatting (3DGS)}~\cite{kerbl20233d}: A neural rendering method based on explicit Gaussian representations.
        \item \textbf{ESLAM}~\cite{johari2023eslam}: An efficient dense SLAM method based on neural implicit representations (employing a hybrid SDF representation with NeRF-style scene encoding).
\end{itemize}

\textbf{Evaluation Metrics.} To comprehensively assess reconstruction quality and system performance, we adopt the following metrics. \textit{In terms of image quality}, we employ three metrics: Peak Signal-to-Noise Ratio (PSNR), SSIM~\cite{wang2004ssim}, and Learned Perceptual Image Patch Similarity (LPIPS)~\cite{zhang2018lpips}. PSNR is defined as:
\begin{equation}
\text{PSNR} = 10 \log_{10}\left(\frac{\text{MAX}^2}{\text{MSE}}\right),
\label{eq:psnr}
\end{equation}
where $\text{MAX}$ represents the maximum pixel value (255), and $\text{MSE} = \frac{1}{N}\sum_{i=1}^{N}(I_i - \hat{I}_i)^2$ is the Mean Squared Error. SSIM~\cite{wang2004ssim} is a perceptual quality index that comprehensively considers luminance, contrast, and structural similarity, with a value range of $[0,1]$ (higher is better). LPIPS~\cite{zhang2018lpips} is a deep feature-based perceptual distance metric, with a value range of $[0,\infty)$ (lower is better).

\textit{Regarding system efficiency}, we utilize three indicators: end-to-end processing time, peak video memory (VRAM) usage, and real-time rendering frame rate.
\subsection{Overall Performance and Baseline Comparison}

\begin{table*}[!t]
\caption{Quantitative comparison on representative regions. \textbf{ParkingTwin} achieves superior quality (highest SSIM/PSNR) and stands as the only method supporting \textbf{real-time processing (30+ FPS)} and \textbf{dynamic object removal} on consumer-grade hardware (GTX 1660).}
\label{tab:comparison}
\centering
{\scriptsize
\setlength{\tabcolsep}{1.5pt}
\renewcommand{\arraystretch}{1.1}
\resizebox{\textwidth}{!}{%
\begin{tabular}{|l|c|c|c|c|c|c|c|c|}
\hline
\multirow{2}{*}{\textbf{Method}} & \multicolumn{3}{c|}{\textbf{Image Quality Metrics}} & \multicolumn{3}{c|}{\textbf{System Efficiency Metrics}} & \multicolumn{2}{c|}{\textbf{Functional Capabilities}} \\
\cline{2-9}
 & \textbf{PSNR}$\uparrow$ & \textbf{SSIM}$\uparrow$ & \textbf{LPIPS}$\downarrow$ & \textbf{Time(min)}$\downarrow$ & \textbf{Mem.(GB)}$\downarrow$ & \textbf{RT} & \textbf{Veh. Rem.} & \textbf{GPU} \\
\hline
3DGS & $26.5 \pm 0.3$ & $0.75 \pm 0.02$ & $0.21 \pm 0.01$ & $74 \pm 5$ & $36.0 \pm 2.0$ & No & No & RTX 4090D \\
\hline
ESLAM & $28.9 \pm 0.4$ & $0.82 \pm 0.03$ & $0.17 \pm 0.02$ & $243 \pm 15$ & $80.0 \pm 5.0$ & No & No & RTX PRO 6000 \\
\hline
\textbf{Ours} & $\textbf{30.1} \pm \textbf{0.2}$ & $\textbf{0.87} \pm \textbf{0.01}$ & $\textbf{0.13} \pm \textbf{0.01}$ & $\textbf{5} \pm \textbf{0.5}$ & $\textbf{6.0} \pm \textbf{0.5}$ & \textbf{Yes} & \textbf{Yes} & \textbf{GTX 1660} \\
\hline
\end{tabular}
}
}
\end{table*}

\begin{figure*}[!t]
        \centering
        \subfloat[Trajectory]{\includegraphics[width=0.24\textwidth]{\detokenize{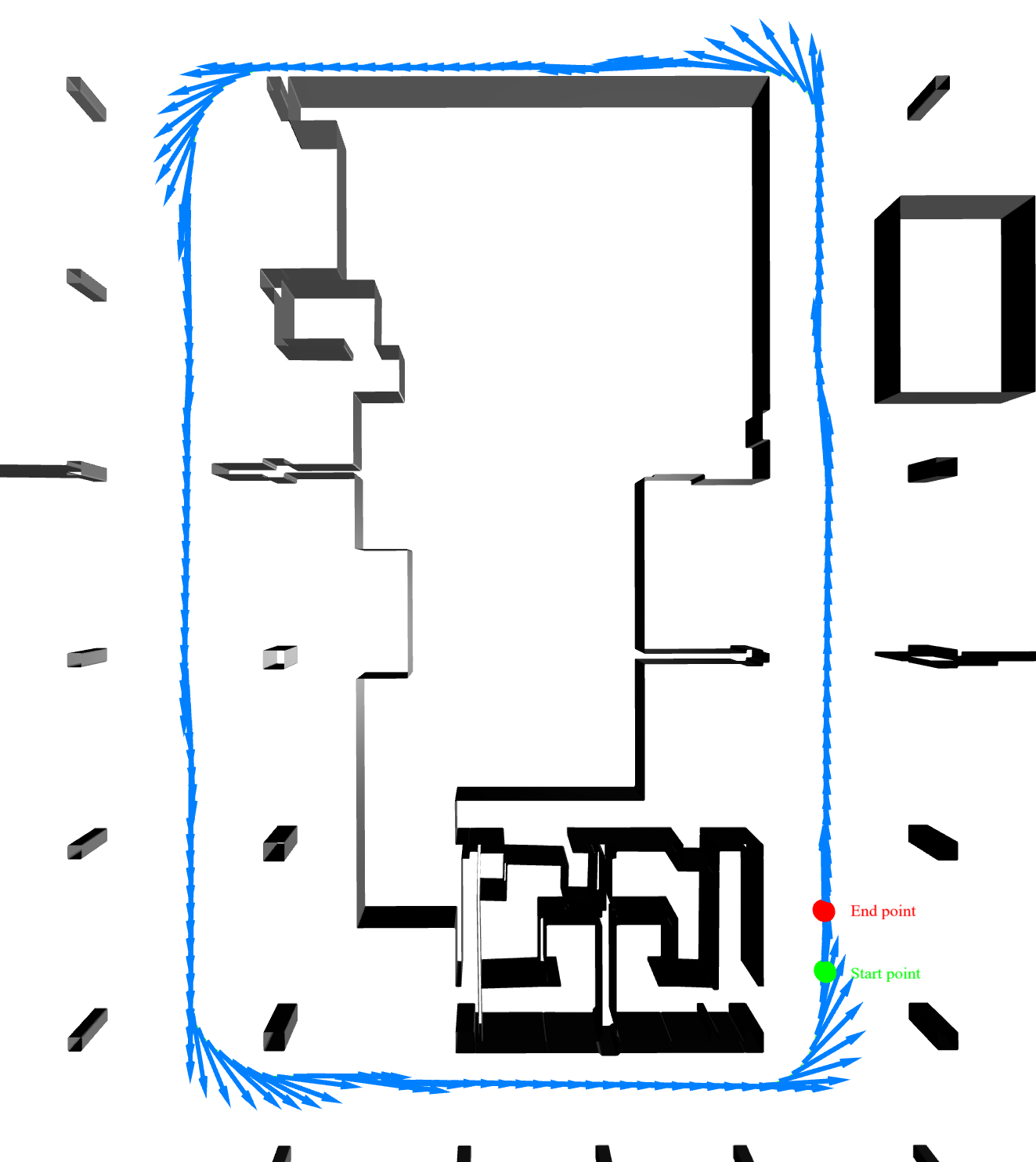}}%
        \label{fig:trajectory}}
        \hfil
        \subfloat[ParkingTwin]{\includegraphics[width=0.21\textwidth]{\detokenize{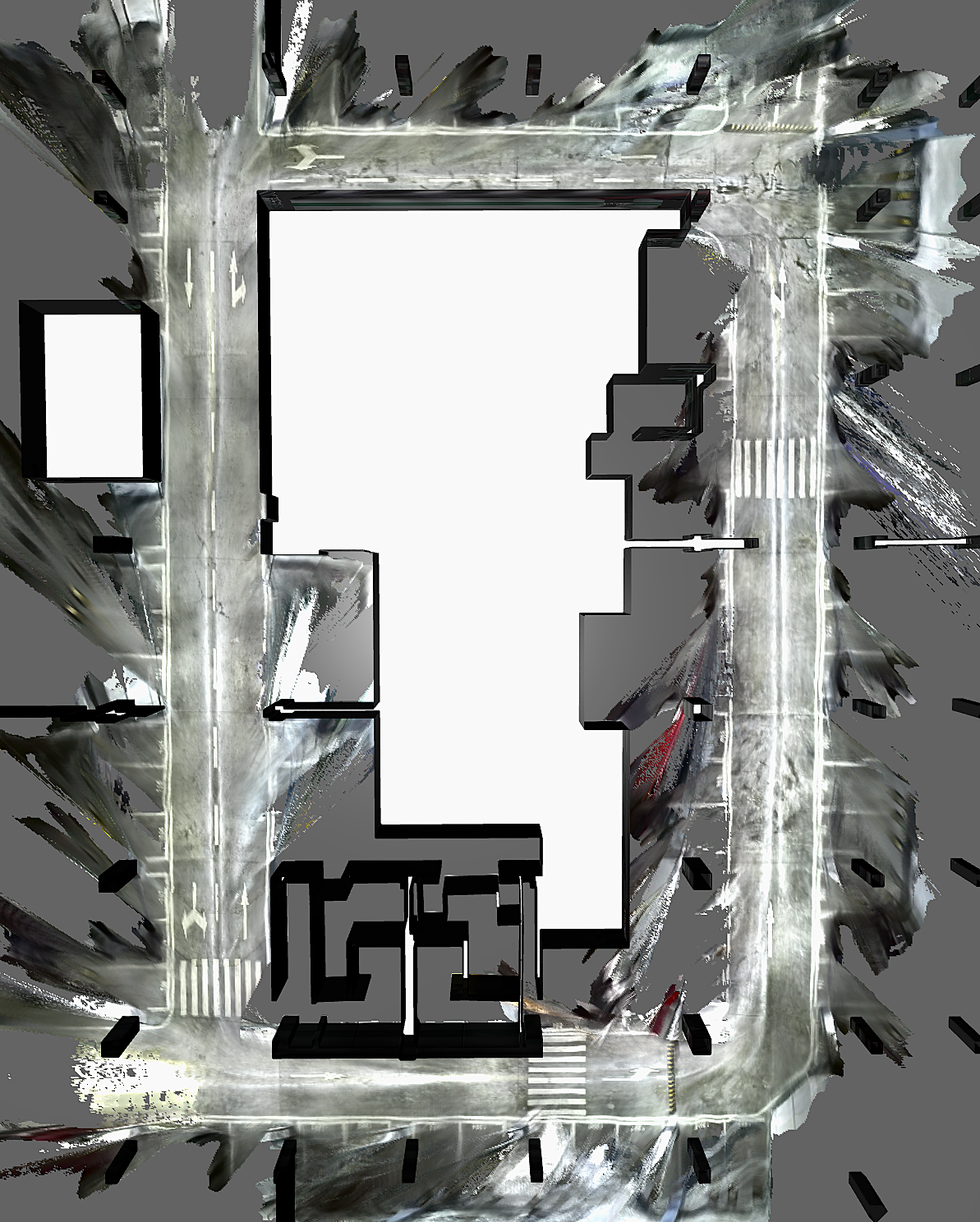}}}
        \hfil
        \subfloat[3DGS]{\includegraphics[width=0.2\textwidth]{\detokenize{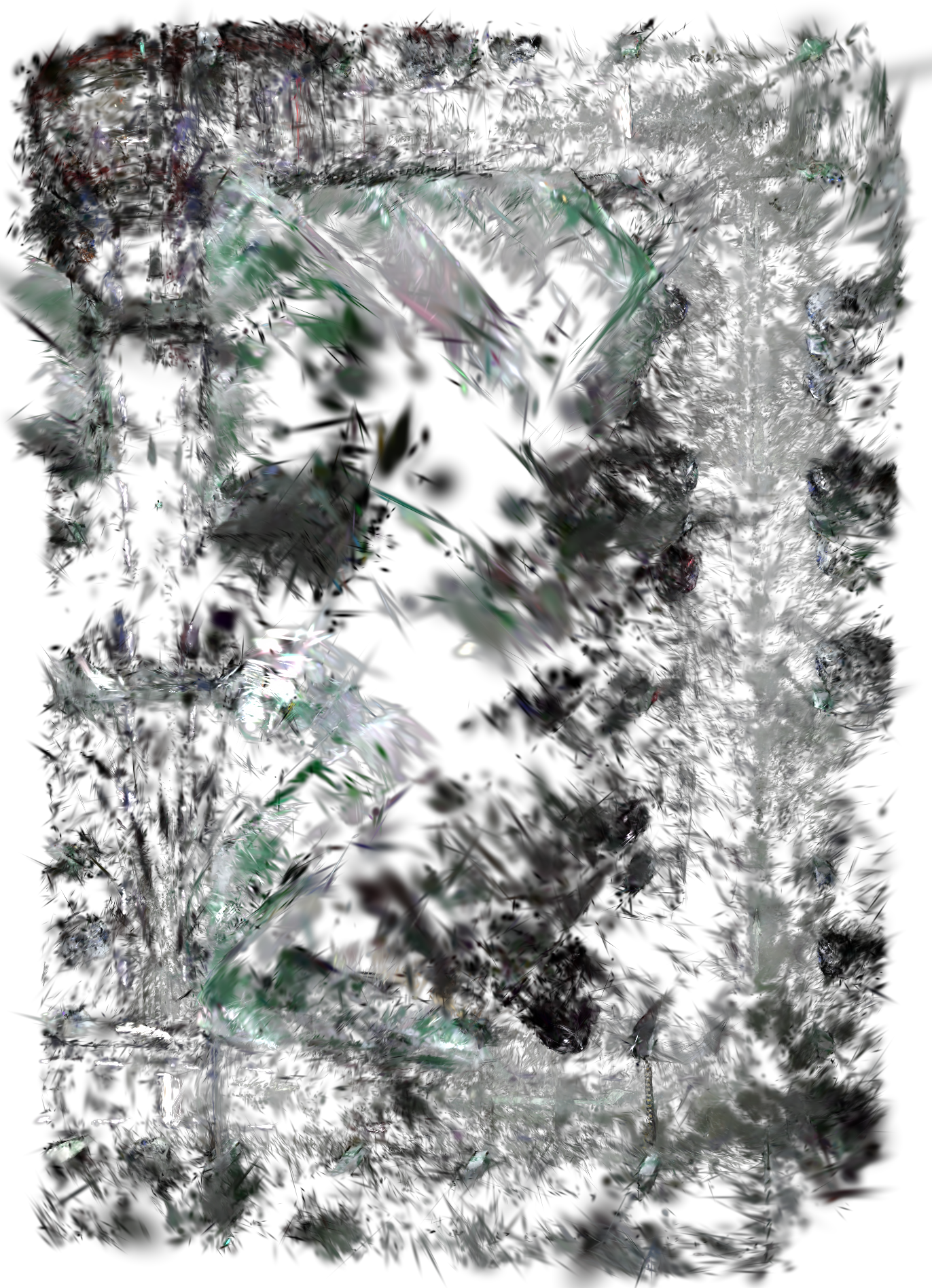}}}
        \hfil
        \subfloat[ESLAM]{\includegraphics[width=0.23\textwidth]{\detokenize{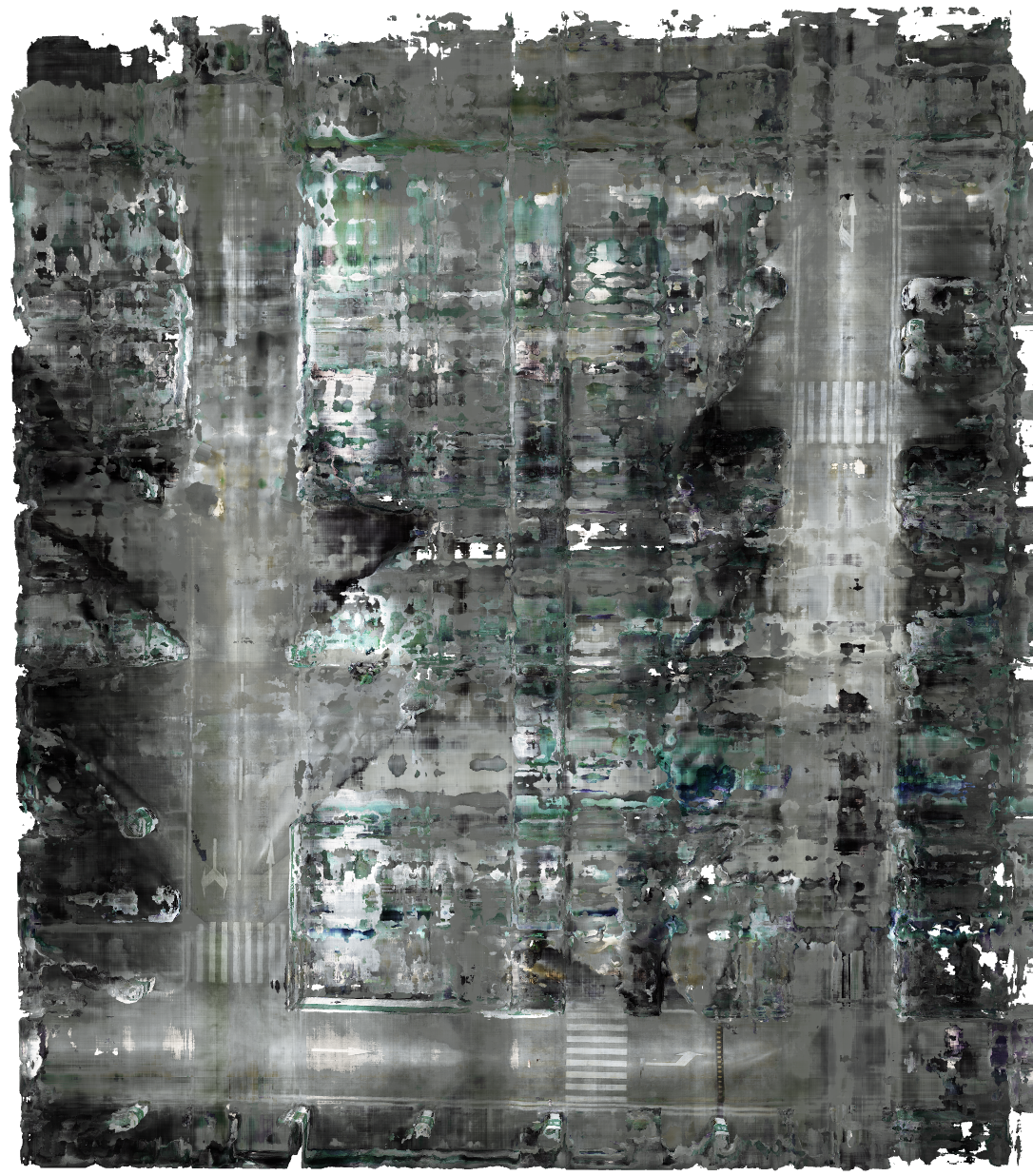}}}
        \caption{Trajectory overview and global reconstruction comparison. (a) The robot executes a typical \textbf{sparse forward-facing inspection path} (blue lines). (b) \textbf{ParkingTwin (Ours)} generates a clean, vehicle-free floor plan. In contrast, (c) \textbf{3DGS} and (d) \textbf{ESLAM} exhibit significant \textbf{ghosting artifacts} and geometric noise due to the lack of structural priors.}
        \label{fig:ours_panorama}
\end{figure*}

\begin{figure*}[!t]
        \centering
        \includegraphics[width=\textwidth]{\detokenize{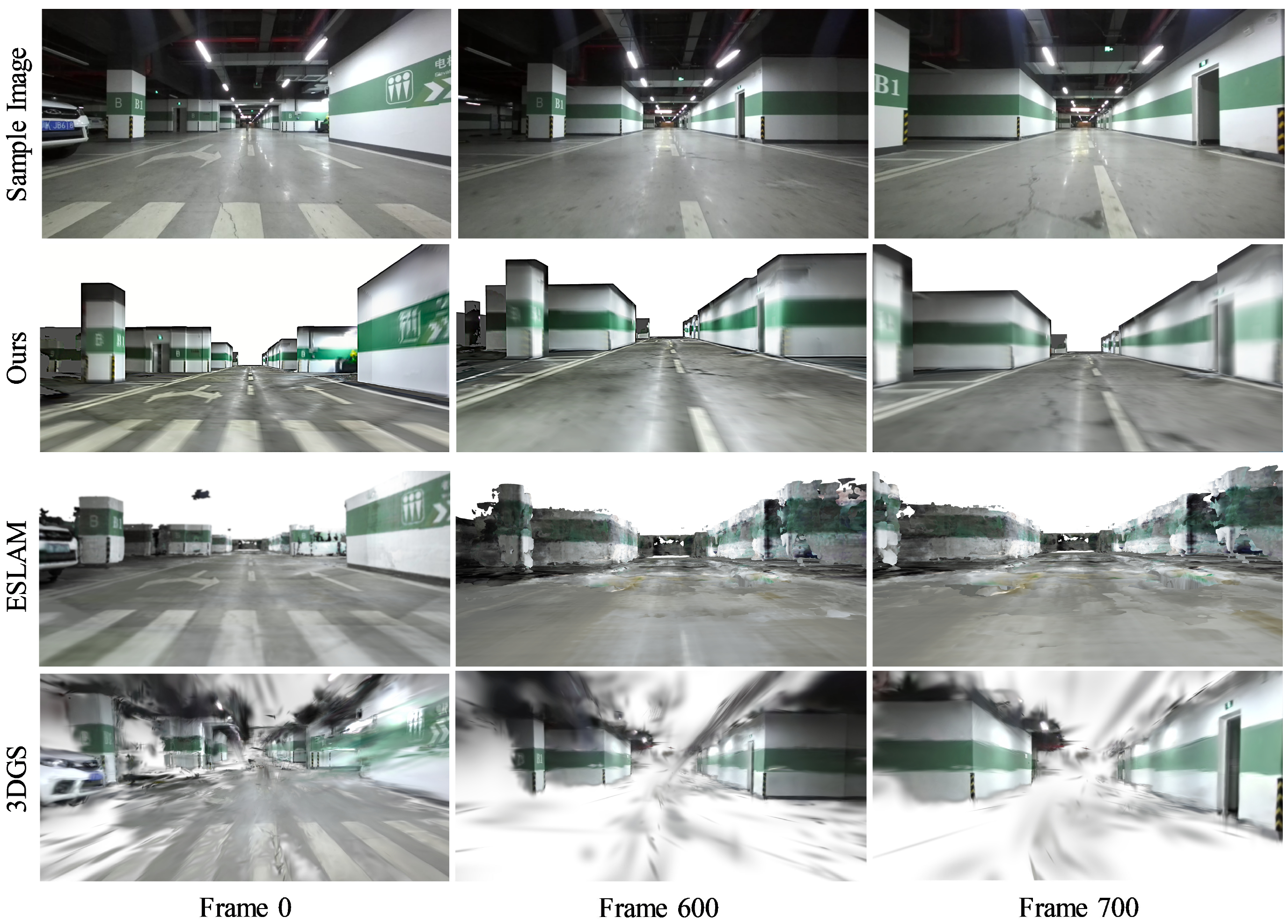}}
        \caption{Detailed comparison of texture quality on keyframes.
        \textbf{Row 1 (Real Input):} Raw captured images containing dynamic vehicle occlusions.
        \textbf{Row 2 (Ours):} ParkingTwin successfully removes dynamic vehicles (Frame 0) and reconstructs clear signage and lane markings (Frames 600/700).
        \textbf{Row 3-4 (Baselines):} ESLAM and 3DGS fail to remove vehicles (resulting in ghosting artifacts) and suffer from severe blurring or \textbf{geometric holes} in textureless regions (e.g., Frame 576).}
        \label{fig:frame_compare}
\end{figure*}

\subsubsection{Quantitative Comparison}
Table~\ref{tab:comparison} presents a comprehensive quantitative comparison with baseline methods in terms of image quality and system efficiency. The experimental design adheres to the following principles: Since the Video Memory (VRAM) complexity of 3DGS scales as $O(N)$ (where $N$ is the number of Gaussians) and ESLAM scales as $O(V)$ (where $V$ is the number of scene voxels), both methods are incapable of processing the entire parking lot (68,000 $\text{m}^2$ per floor) due to resource exhaustion. To validate the advantage of prior information in breaking computational resource bottlenecks, we selected five representative sub-regions (each containing approximately 4,300 frames). All methods were executed independently five times to ensure statistical reliability. In contrast, ParkingTwin exhibits an $O(1)$ VRAM complexity (dependent solely on single-frame resolution, independent of scene scale), enabling scalable processing of scenarios of arbitrary magnitude. Additionally, OpenMVS~\cite{cernea2015openmvs} was excluded from the quantitative comparison. As it relies on multi-angle views for dense stereo matching, the insufficient parallax inherent in sparse forward-facing views led to catastrophic reconstruction failure (see Fig.~\ref{fig:openmvs_failure}).

\noindent\textbf{Image Quality Metrics.} ParkingTwin achieves an SSIM of $0.87 \pm 0.01$, significantly outperforming 3DGS ($0.75 \pm 0.02$, a relative improvement of 16.0\%) and ESLAM ($0.82 \pm 0.03$, +6.1\%). Regarding PSNR, our method attains $30.1 \pm 0.2$ dB, surpassing 3DGS ($26.5 \pm 0.3$ dB, +3.6 dB) and ESLAM ($28.9 \pm 0.4$ dB, +1.2 dB). For LPIPS, our score is $0.13 \pm 0.01$, which is lower than that of ESLAM ($0.17 \pm 0.02$, a reduction of 23.5\%) and 3DGS ($0.21 \pm 0.01$, -38.1\%). Notably, ParkingTwin yields optimal results across all image quality metrics with the minimal standard deviation (e.g., SSIM std is only 0.01), demonstrating the method's superior stability and robustness. Furthermore, the ability of ParkingTwin to generate clean, vehicle-free textures is crucial for downstream parking lot digital twin applications.

\noindent\textbf{System Efficiency Metrics.} ParkingTwin exhibits substantial efficiency advantages. The end-to-end processing time is merely $5 \pm 0.5$ minutes, yielding approximately $15\times$ speedup over 3DGS ($74 \pm 5$ minutes) and $49\times$ speedup over ESLAM ($243 \pm 15$ minutes). The complete 4,300-frame processing workflow encompasses OSM geometric initialization, vehicle detection, texture projection and fusion, seam detection and smoothing, and mesh export. Peak Video Memory (VRAM) usage is limited to $6.0 \pm 0.5$ GB, which is drastically lower than that of ESLAM ($80.0 \pm 5.0$ GB, a reduction of 92.5\%) and 3DGS ($36.0 \pm 2.0$ GB, -83.3\%). Regarding hardware requirements, ParkingTwin is fully operable on an entry-level GTX 1660 GPU. In stark contrast, 3DGS requires a high-end RTX 4090D, and ESLAM demands a professional-grade Quadro RTX 6000. This demonstrates the exceptional hardware compatibility and deployment flexibility of our system.

\noindent\textbf{Dynamic Cleaning Capability.} This stands as a distinctive advantage of ParkingTwin: among all compared methods, ours is the only one capable of automatically removing vehicles to generate an occlusion-free, clean environmental model. In contrast, all baseline methods permanently ``bake'' dynamic vehicles into the model: 3DGS solidifies vehicles as part of the scene geometry during optimization, while ESLAM's neural field encoding similarly fails to purge transient occlusions after training. This capability is indispensable for parking lot digital twin applications, which necessitate a vehicle-free environment for autonomous driving simulation and path planning. Quantitative experiments indicate that the SSIM in occlusion-affected regions improves from 0.72 to 0.88 (a relative gain of 22.2\%) after vehicle removal, effectively recovering the intrinsic texture of the scene.

\subsubsection{Qualitative Comparison}
Figures~\ref{fig:ours_panorama} and \ref{fig:frame_compare} illustrate the qualitative comparison results in representative regions. \textbf{Figure~\ref{fig:ours_panorama}(a) depicts the data acquisition trajectory}, showing the robot moving unidirectionally along the driving lane to form a closed-loop path, which aligns with the constraints of actual inspection scenarios. Figures~\ref{fig:ours_panorama}(b)-(d) compare the reconstruction results of the three methods based on this trajectory data. Figure~\ref{fig:frame_compare} further presents a detailed comparison of texture quality at the keyframe level. We conduct a detailed analysis from the following four dimensions:

\noindent\textbf{(1) Dynamic Vehicle Handling Capability.} In Frame 0, a white vehicle is present on the left side of the real input image. The reconstruction results reveal that: ParkingTwin successfully removes the vehicle, restoring the ground texture of the parking space; 3DGS retains blurry ghosting of the vehicle and produces severe ``oil-painting-like'' smearing artifacts in the vehicle area; ESLAM similarly retains the vehicle, surrounded by distinct black geometric voids. This comparison validates the effectiveness of our multi-modal geometric detection method.

\noindent\textbf{(2) Reconstruction Quality of Critical Navigation Signage.} For navigation signage crucial to autonomous driving simulation (e.g., lane markings, zone indicator ``B1'', elevator signs), the three methods exhibit significant performance disparities. In Frame 0, located near the starting point (covered by multi-view overlap), ParkingTwin clearly reconstructs the cylindrical ``B''/``B1'' signs and the elevator icon on the right; 3DGS suffers from severe degradation of all signs due to global motion blur, appearing as overexposed whitish diffusion; ESLAM renders identifiable signs but with blurred edges. As the robot moves to the distal ends of the trajectory in Frames 0576 and 0676 (sparse view coverage), the data sparsity issue intensifies: ground lane markings in 3DGS vanish completely, with the entire floor appearing as a uniform grayish-white surface; ESLAM exhibits extensive black geometric voids on walls and yellow-green chromatic noise contamination on the ground; in contrast, ParkingTwin maintains identifiable lane markings and clear wall striations.

\noindent\textbf{(3) Geometric Integrity and Texture Consistency.} Under sparse views, ESLAM suffers from massive geometric voids due to the lack of multi-view supervision. Although 3DGS maintains geometric completeness, accurate Gaussian covariance estimation becomes infeasible under sparse observations, leading to texture blurring and floating artifacts. This indicates that \textit{blind reconstruction} struggles to reliably parse geometry under sparse views, where geometric ambiguity is an inevitable consequence of the ill-posed nature of the problem. ParkingTwin anchors texture fusion onto the OSM-driven TSDF geometry, circumventing the geometric optimization bottleneck and maintaining stable reconstruction quality even under sparse views.

\begin{figure}[!t]
        \centering
        \includegraphics[width=0.6\columnwidth]{\detokenize{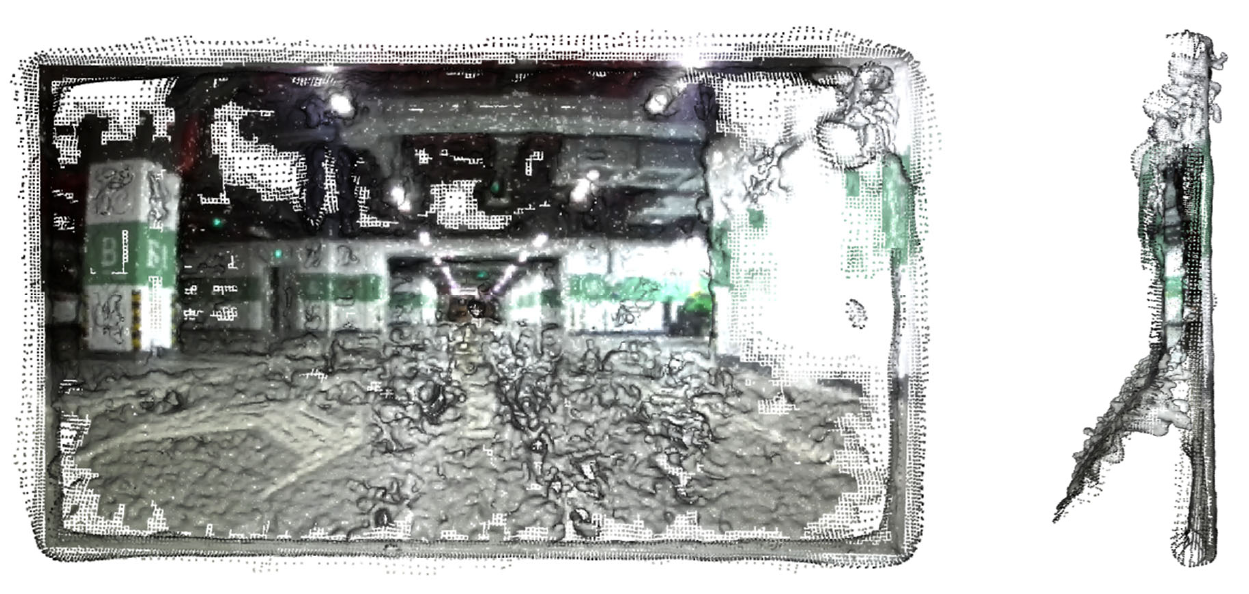}}
        \caption{Failure of Traditional MVS (OpenMVS) under sparse views. Due to the lack of multi-view parallax, dense matching becomes mathematically \textbf{ill-posed}, resulting in >60\% geometric loss and planar compression. This fundamental physical constraint validates the necessity of our \textbf{OSM-driven prior}.}
        \label{fig:openmvs_failure}
\end{figure}

\noindent\textbf{(4) Failure Case of Traditional MVS Methods.} Figure~\ref{fig:openmvs_failure} demonstrates that OpenMVS exhibits holes in approximately 60--70\% of the area under sparse forward-facing views, with the point cloud compressed onto a plane. This failure stems from \textit{physical constraints}: insufficient parallax renders MVS inherently ill-posed, rather than reflecting an algorithmic defect. This observation further confirms that, under sparse views, incorporates external geometric priors is a necessary condition to transform reconstruction from an ill-posed domain to a solvable domain.

\subsection{Quality Assessment of OSM Geometric Initialization}

\begin{figure}[!t]
\centering
\subfloat[OSM Vector Map]{\includegraphics[width=0.45\columnwidth]{\detokenize{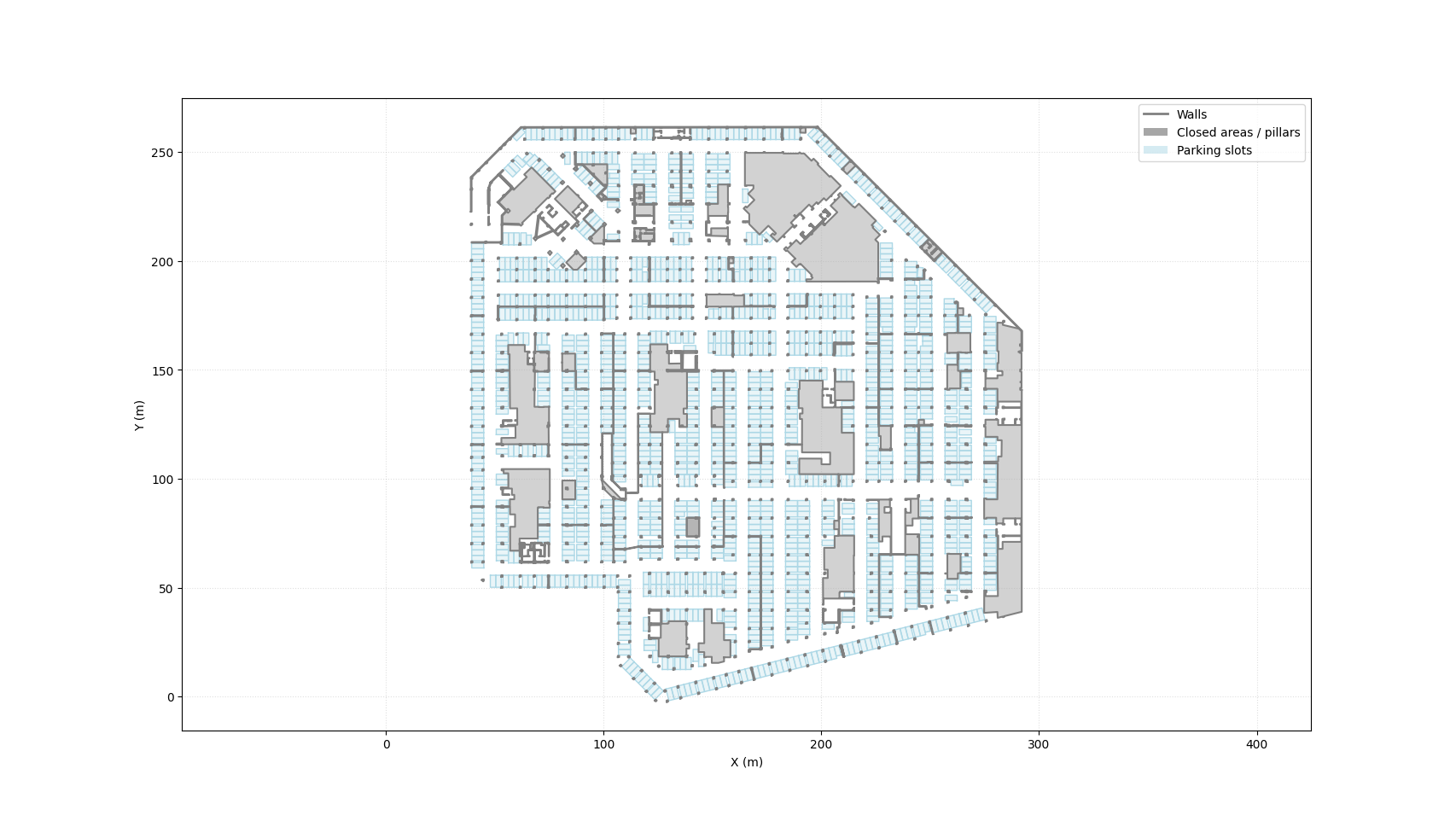}}%
\label{fig:osm}}
\hfil
\subfloat[TSDF Triangular Mesh]{\includegraphics[width=0.45\columnwidth]{\detokenize{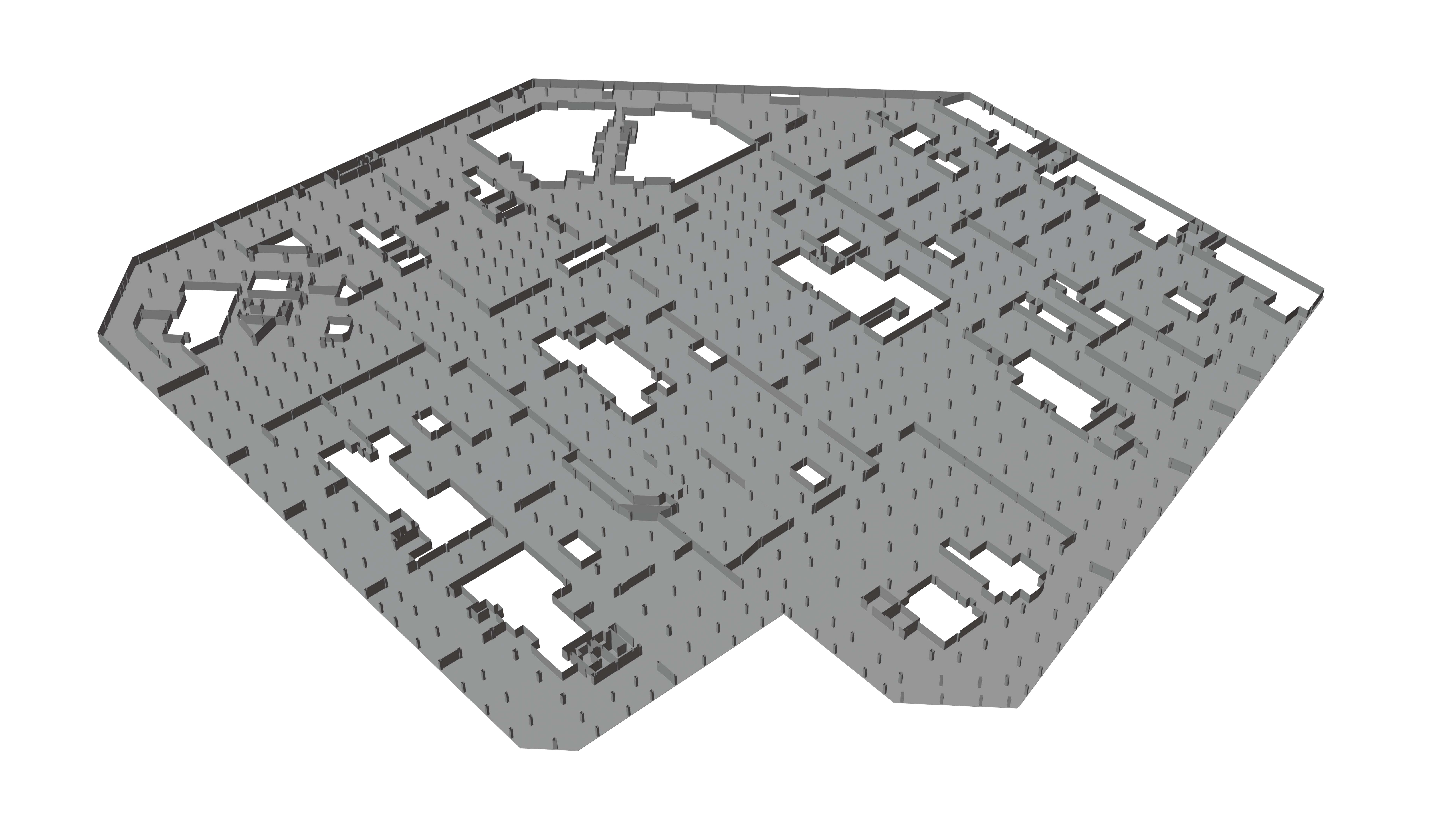}}%
\label{fig:tsdf}}
\caption{Process of OSM-driven geometric initialization. (a) OSM vector map. (b) The generated 3D TSDF mesh exhibits \textbf{clean topology} and \textbf{manifold geometry}, avoiding the noise typical of depth fusion methods.}
\label{fig:osm_tsdf}
\end{figure}

Figure~\ref{fig:osm_tsdf} illustrates the OSM map and the generated TSDF mesh. The OSM map (Fig.~\ref{fig:osm_tsdf}(a)) was generated by our team by extracting geometric structures from the parking lot's CAD blueprints and converting them\footnote{Our team developed an AutoCAD-to-OSM conversion tool. The map generation results and commercial applications can be viewed at: \url{https://www.daodajiao.com/ditu.html}.}~\cite{cao2025osm2net}, followed by precise alignment with the real-world environment through LiDAR point cloud registration~\cite{qiu2025lidar}. The registered OSM map is then converted into a topologically clean triangular mesh (Fig.~\ref{fig:osm_tsdf}(b)) via our 3D TSDF generation algorithm.

We validate the quality of OSM-driven TSDF initialization from two perspectives. \textit{In terms of metric accuracy}: We compared the generated TSDF mesh with actual measurements of key scene dimensions (column spacing, parking space width, lane width). The mean relative error is 1.8\% with a standard deviation of 0.9\%, validating the accuracy of the metric information provided by OSM. \textit{In terms of topological quality}: Compared to meshes reconstructed \textit{from scratch} by traditional MVS methods (e.g., OpenMVS), which typically suffer from significant geometric noise, holes, and topological errors, the mesh generated by our OSM-prior driven method is topologically clean, hole-free, and free of self-intersecting faces. This provides a stable geometric foundation for subsequent texture mapping. This advantage stems directly from the precise geometric information of the CAD blueprints and the high-precision alignment via LiDAR registration.

\subsection{Ablation Study}

This section validates the effectiveness and necessity of the core technical modules of ParkingTwin through a systematic ablation study. We adopt a strict evaluation protocol: the full 4,300 frames are utilized for reconstruction, while image quality metrics are computed on a 500-frame evaluation set. This set is uniformly sampled from the complete sequence to comprehensively cover diverse scenarios, including illuminated areas, dark zones, lighting transition regions, and dynamic occlusion cases. Each experimental configuration is executed independently three times, with results reported as Mean $\pm$ Standard Deviation.

\subsubsection{Effectiveness of the OSM Geometric Prior}
\label{sec:ablation_osm}

\begin{table*}[!t]
\caption{Ablation Study: Depth Fusion vs. OSM Prior. The OSM prior not only elevates structural completeness to \textbf{100\%} but also reduces the vertex count by \textbf{93.3\%} by eliminating noise, thereby enabling efficient real-time rendering.}
\label{tab:osm_ablation}
\centering
{\scriptsize
\setlength{\tabcolsep}{1.5pt}
\renewcommand{\arraystretch}{1.1}
\resizebox{\textwidth}{!}{%
\begin{tabular}{|l|c|c|c|c|c|c|}
\hline
\textbf{Geometry initialization method} & \textbf{PSNR}$\uparrow$ & \textbf{SSIM}$\uparrow$ & \textbf{Comp. (\%)}$\uparrow$ & \textbf{\#Verts.} & \textbf{\#Faces} & \textbf{Time(s)} \\
\hline
Depth fusion (no prior) & 25.8 & 0.69 & 72\% & 3.293M & 5.823M & 333.4 \\
\textbf{OSM prior (ours)} & \textbf{30.1} & \textbf{0.87} & \textbf{100\%} & \textbf{0.22M} & \textbf{0.42M} & \textbf{8.0} \\
\hline
\textit{Relative improvement} & \textit{+4.3 dB} & \textit{+26.1\%} & \textit{+28\%} & \textit{-93.3\%} & \textit{-92.8\%} & \textit{-97.6\%} \\
\hline
\end{tabular}
}
}
\end{table*}

\begin{figure}[!t]
\centering
\subfloat[Depth Fusion (no prior)]{\includegraphics[width=0.3\textwidth]{\detokenize{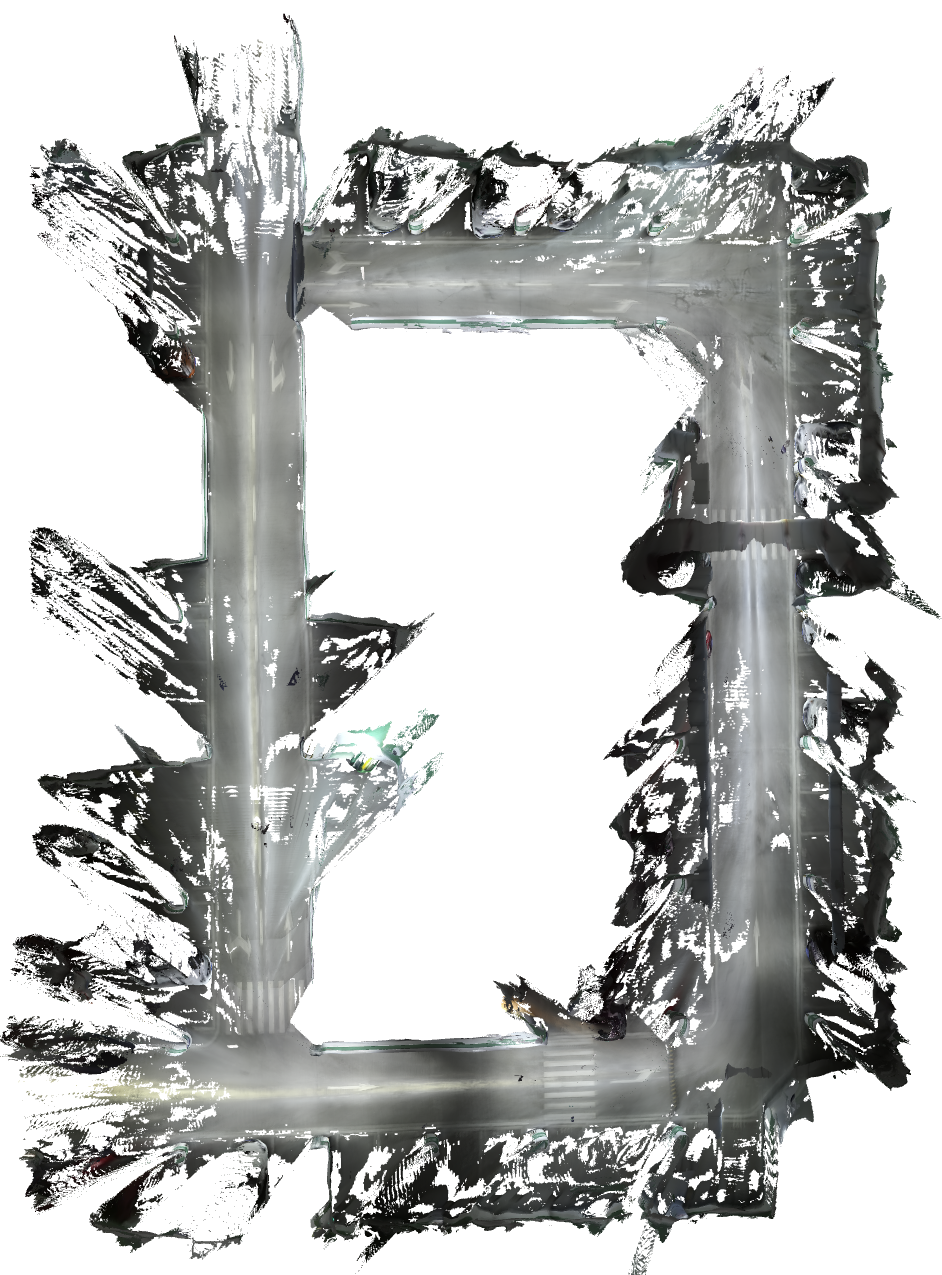}}%
\label{fig:ablation_depth_fusion}}
\hfil
\subfloat[OSM Prior (ours)]{\includegraphics[width=0.3\textwidth]{\detokenize{ours_birdseye.png}}%
\label{fig:ablation_osm_prior}}
\caption{Visual comparison of geometric initialization strategies. (a) \textbf{Depth Fusion (no prior)} suffers from noise, jagged boundaries, and internal distortion due to sensor noise. (b) \textbf{OSM Prior (ours)} ensures straight walls, uniform column spacing, and clean topology consistent with architectural standards.}
\label{fig:ablation_geometry_comparison}
\end{figure}

In this section, we validate the contribution of the OSM prior to geometric initialization through a controlled ablation study, while keeping other system components fixed. Table~\ref{tab:osm_ablation} and Fig.~\ref{fig:ablation_geometry_comparison} present a comparative analysis of the performance of two geometric initialization strategies on representative regions.

The \textbf{Depth Fusion} method directly integrates RGB-D depth maps to generate meshes. However, the lack of structural constraints results in significant geometric noise (completeness of 72\%). It requires 333.4 seconds to process 4,302 frames, generating 3.293 million vertices, a large portion of which are located in invalid regions. In contrast, the \textbf{OSM Prior} leverages semantic structures to generate the TSDF, achieving 100\% completeness in just 8 seconds (a \textbf{41.7$\times$} acceleration). Although it generates only 220k vertices, \textit{every vertex corresponds to an effective scene structure}, ensuring the topology aligns with design specifications.

Quantitative results indicate that compared to the depth fusion method, the OSM prior elevates SSIM from 0.69 to 0.87 (+26.1\%), increases PSNR by 4.3 dB, and improves completeness by 28\%. Although the depth fusion method yields a higher vertex count, its \textbf{effective geometric density} is merely 15\% of that achieved by the OSM method, as massive vertices are wasted in noisy regions outside the boundaries. Conversely, the OSM method achieves \textbf{design-spec level topological accuracy}, providing a stable geometric foundation for subsequent texture mapping. This ablation demonstrates that the OSM semantic prior not only provides a topologically clean geometric framework but also achieves \textbf{order-of-magnitude computational acceleration}, enabling ParkingTwin to attain high-quality reconstruction even under sparse views.

\begin{figure}[!t]
        \centering
        \subfloat[w/o Gradient Weight]{\includegraphics[width=0.4\columnwidth]{\detokenize{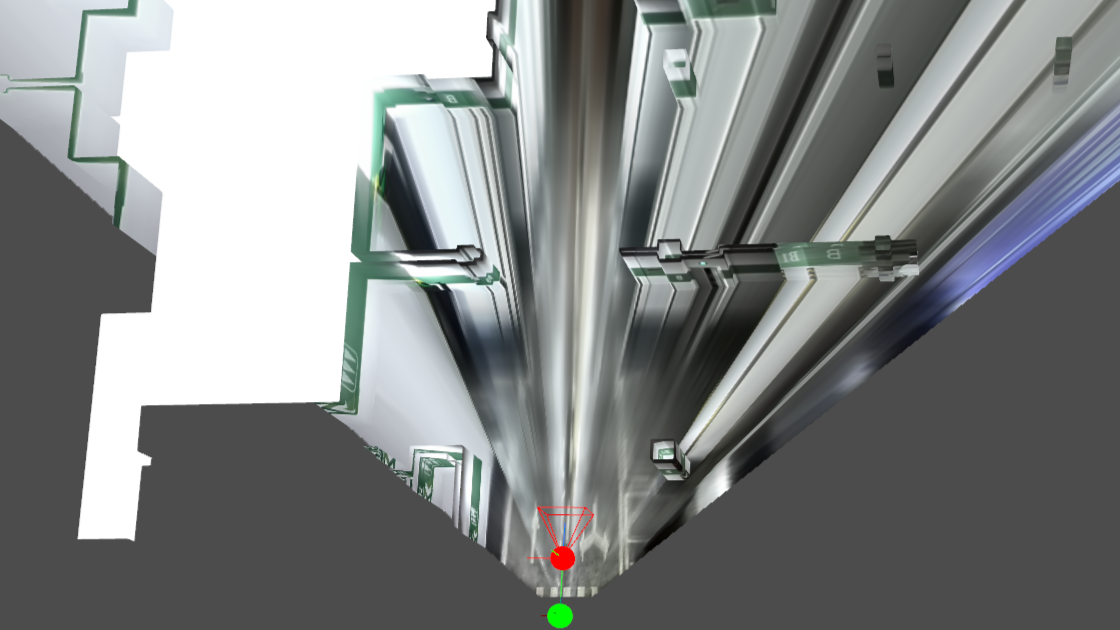}}%
        \label{fig:depth_no}}
        \hfil
        \subfloat[w/ Gradient Weight]{\includegraphics[width=0.4\columnwidth]{\detokenize{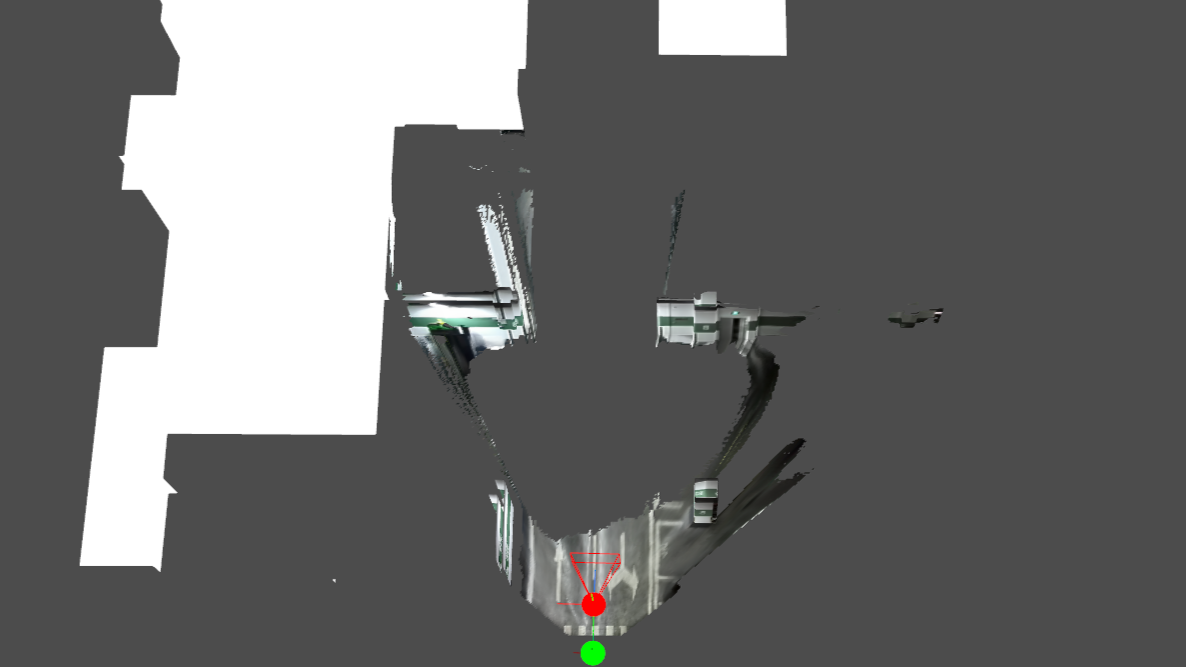}}%
        \label{fig:depth_yes}}
        \caption{Ablation study on the edge-preserving effect of the depth gradient weight $w_g$. (a) Without $w_g$, significant \textbf{color bleeding} and artifacts appear at occlusion boundaries (e.g., column edges); (b) With $w_g$ enabled, erroneous projections are effectively suppressed, preserving the \textbf{sharpness} of texture edges.}
        \label{fig:depth_consistency}
        \end{figure}
        
        \begin{figure*}[!t]
        \centering
        \subfloat[Baseline]{\includegraphics[width=0.3\columnwidth]{\detokenize{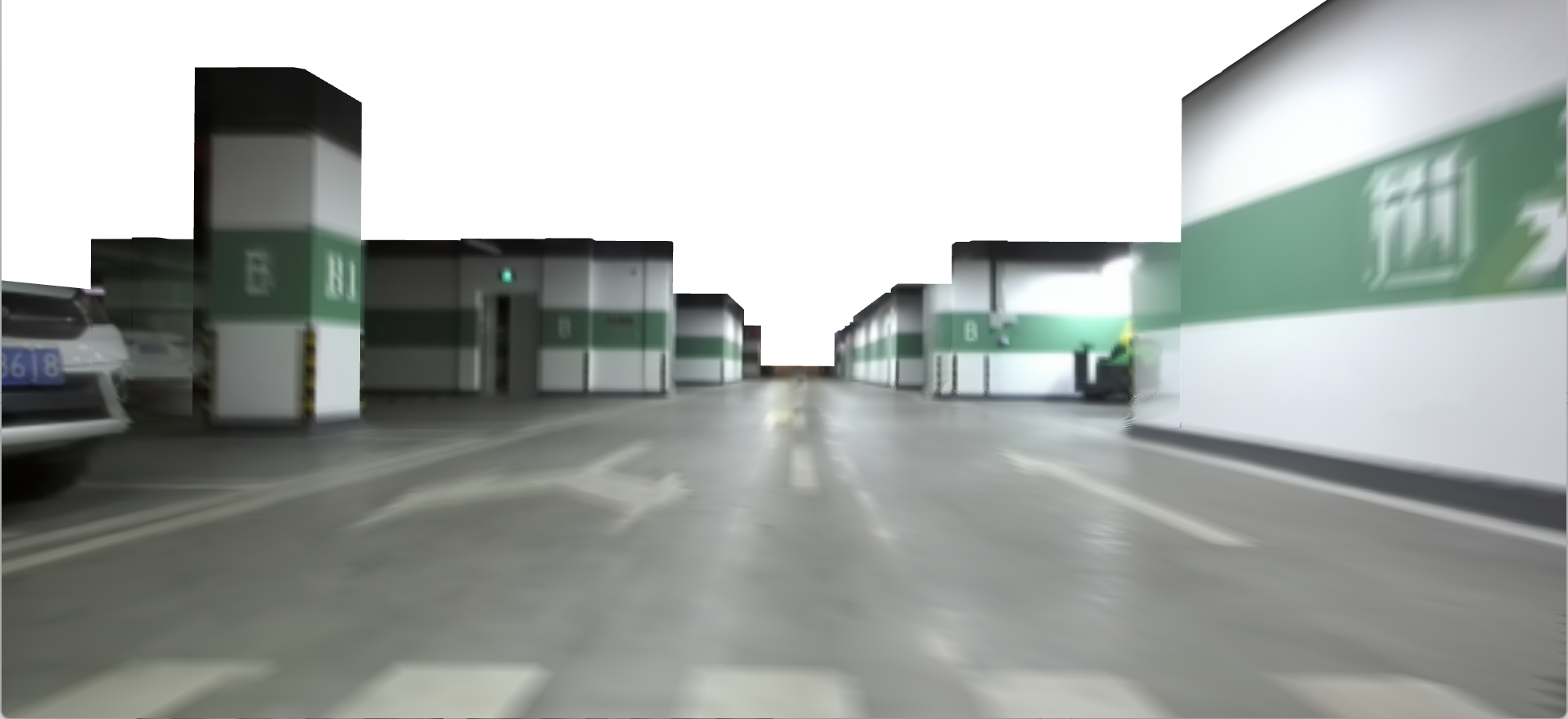}}%
        \label{fig:module_cum_local_a}}
        \hfil
        \subfloat[+ Veh. Rem.]{\includegraphics[width=0.3\columnwidth]{\detokenize{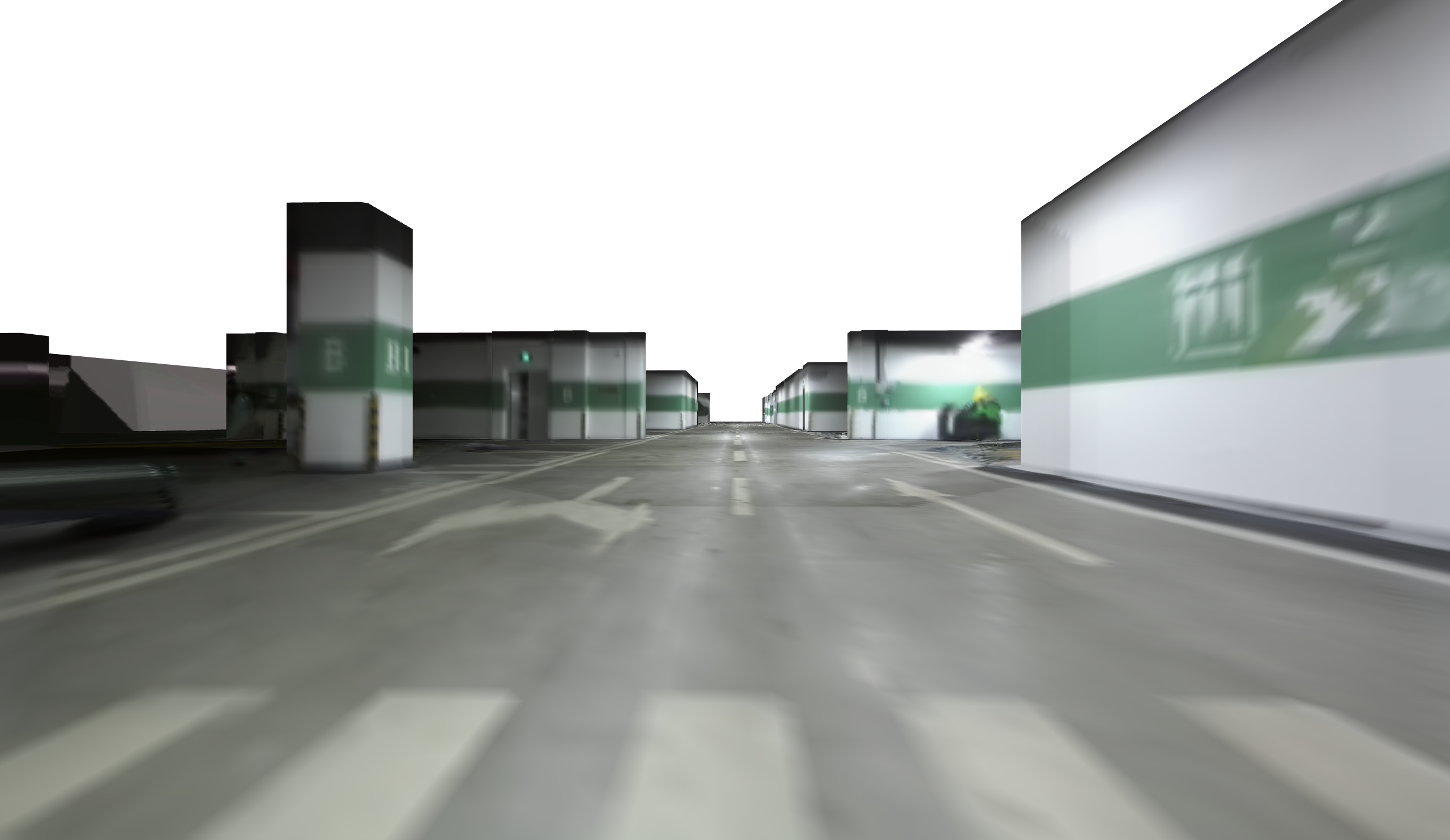}}%
        \label{fig:module_cum_local_c}}
        \hfil
        \subfloat[+ LAB Fusion]{\includegraphics[width=0.3\columnwidth]{\detokenize{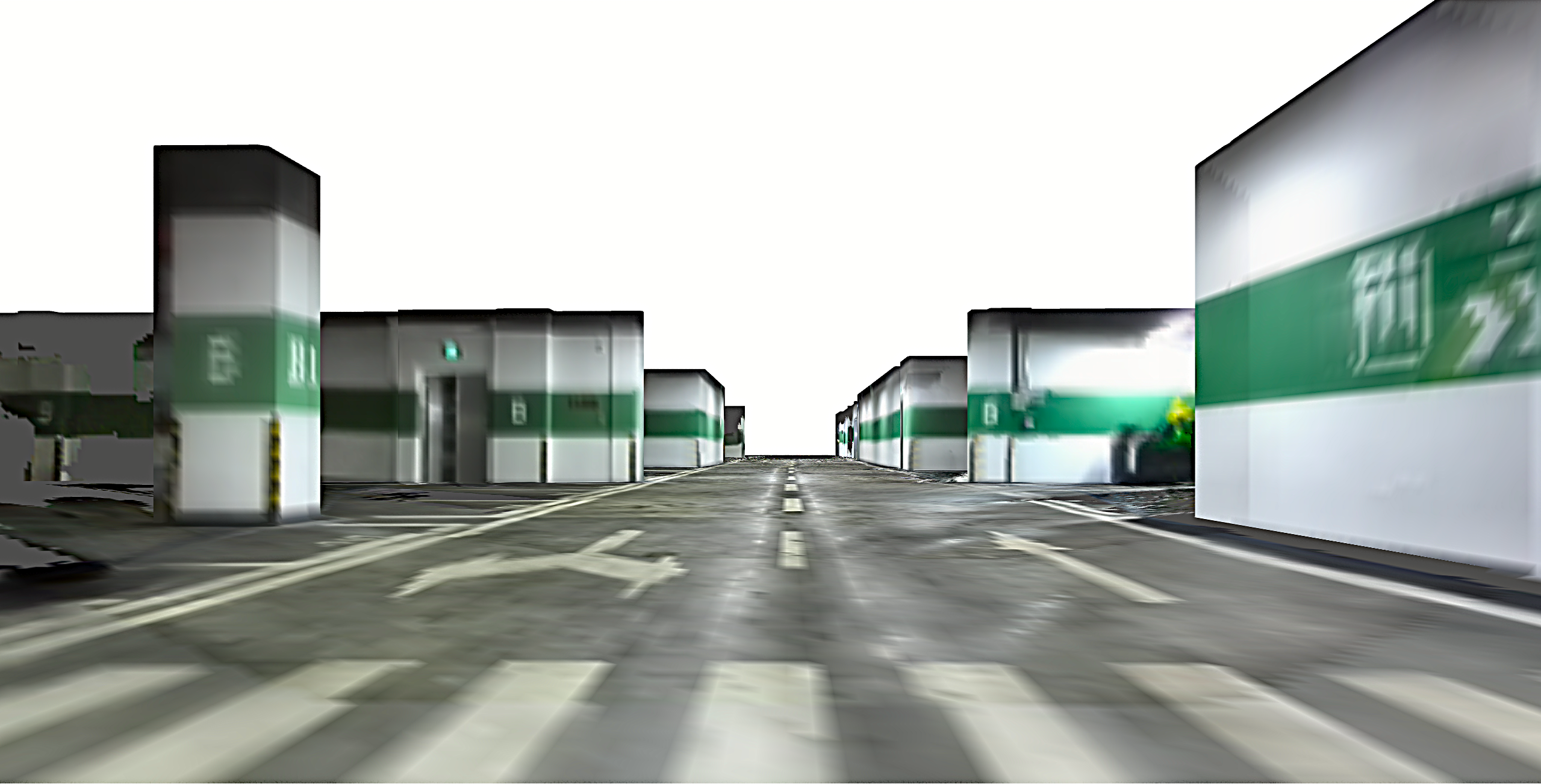}}%
        \label{fig:module_cum_local_d}}
        \caption{Progressive ablation study of core modules (local close-up view). (a) \textbf{Baseline (RGB):} Exhibits significant \textbf{vehicle ghosting} and illumination inconsistency; (b) \textbf{+ Dyn. Filter:} Vehicles are successfully removed, yet \textbf{blocky seams} caused by abrupt lighting changes persist on the road surface; (c) \textbf{Full System (LAB):} With the introduction of LAB perceptual fusion, seams are completely eliminated, yielding a continuous and uniform road surface texture.}
        \label{fig:module_cum_local}
        \end{figure*}
        
        \begin{figure*}[!t]
        \centering
        \subfloat[Baseline]{\includegraphics[width=0.3\columnwidth]{\detokenize{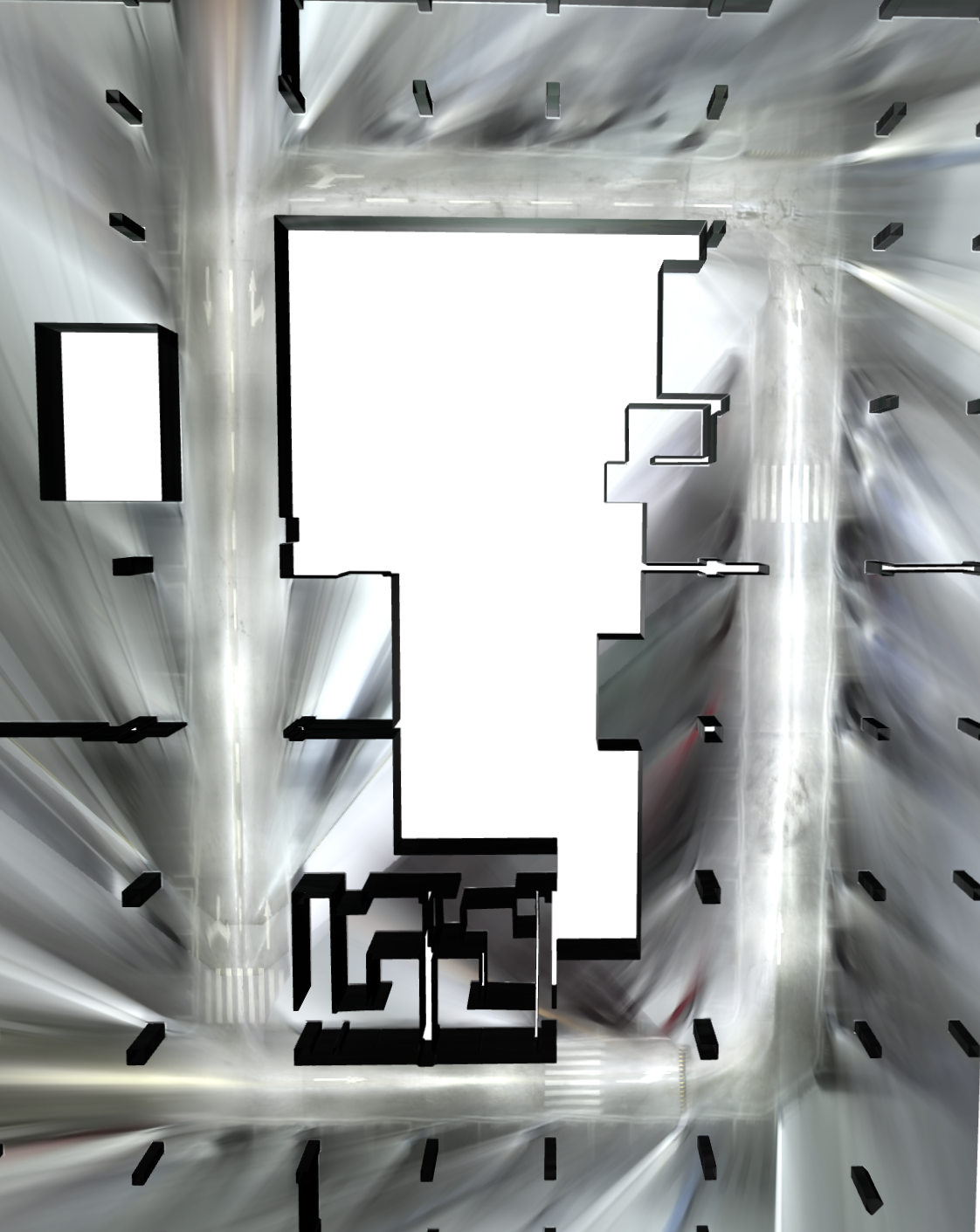}}%
        \label{fig:module_cum_global_a}}
        \hfil
        \subfloat[+ Veh. Rem.]{\includegraphics[width=0.31\columnwidth]{\detokenize{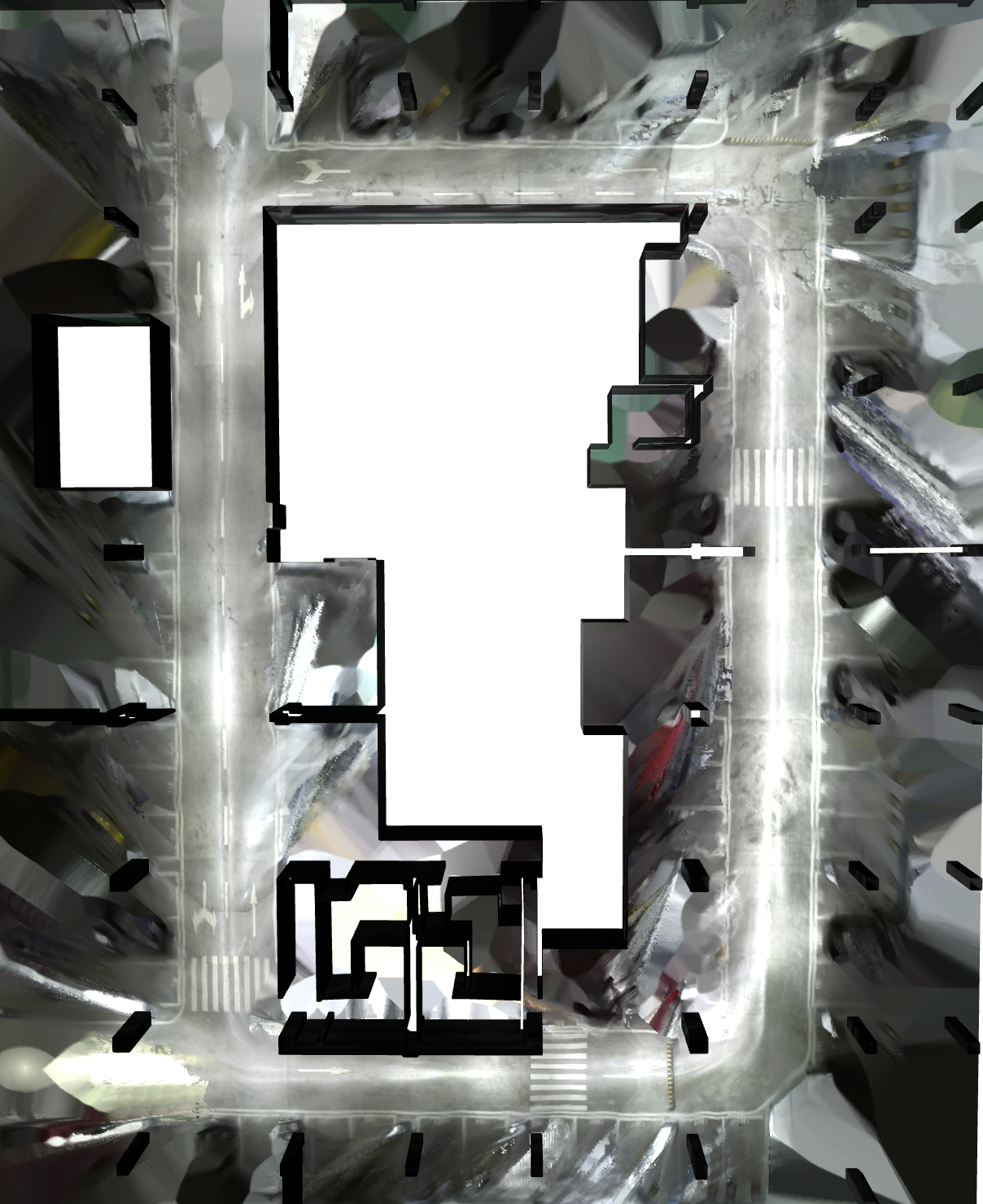}}%
        \label{fig:module_cum_global_c}}
        \hfil
        \subfloat[Full Sys.]{\includegraphics[width=0.31\columnwidth]{\detokenize{ours_birdseye.png}}%
        \label{fig:module_cum_global_d}}
        \caption{Progressive ablation study of core modules (global top-down view). Compared to the local view, the global perspective more clearly reveals the consistency of ParkingTwin across the large-scale scene. In (b), although dynamic vehicles are removed, the ground surface still exhibits a \textbf{``patchy''} appearance; (c) demonstrates that the LAB fusion strategy successfully achieves \textbf{global color balance}.}
        \label{fig:module_cum_global}
        \end{figure*}

\subsubsection{Analysis of Cumulative Module Contributions}
\label{sec:ablation_cumulative}
        
In this section, we evaluate the cumulative contribution of each module to the overall performance by \textbf{progressively enabling core components in conjunction with visual results}. The experiment examines three representative configurations (see Table~\ref{tab:ablation_cumulative}): Cfg.~A serves as the \textit{Baseline}, performing simple weighted fusion exclusively in RGB space; Cfg.~B incorporates geometry-driven vehicle detection and occlusion filtering into the RGB pipeline but without LAB texture fusion; Cfg.~C represents the \textit{Full System}, which enables LAB perceptual space fusion, depth gradient weighting, and seam smoothing upon the foundation of Cfg.~B. Note that the Geometric Initialization Module $\mathcal{M}_\text{geo}$ remains active throughout this subsection, as its independent contribution has already been analyzed in Sec.~\ref{sec:ablation_osm}; consequently, this analysis focuses primarily on the incremental impact of $\mathcal{M}_\text{dyn}$ and $\mathcal{M}_\text{tex}$ on \textbf{texture quality}. Quantitative results are presented in Table~\ref{tab:ablation_cumulative}, with corresponding \textbf{local close-up views} shown in Fig.~\ref{fig:module_cum_local} and \textbf{global top-down views} in Fig.~\ref{fig:module_cum_global}.

\begin{table*}[!t]
\caption{Cumulative contribution of core modules. The progressive integration of \textbf{vehicle detection} (Cfg.~B) and \textbf{LAB fusion} (Cfg.~C) consistently improves PSNR and SSIM, validating the effectiveness of each component.}
\label{tab:ablation_cumulative}
\centering
{\scriptsize
\setlength{\tabcolsep}{1.5pt}
\renewcommand{\arraystretch}{1.1}
\resizebox{\textwidth}{!}{%
\begin{tabular}{|l|c|c|c|c|}
\hline
\textbf{Cfg.} & \textbf{PSNR}$\uparrow$ & \textbf{SSIM}$\uparrow$ & \textbf{LPIPS}$\downarrow$ & \textbf{$\Delta$SSIM} \\
\hline
Cfg. A: Baseline (RGB fusion) & 26.5 & 0.72 & 0.24 & -- \\
Cfg. B: + vehicle detection (no LAB) & 29.1 & 0.82 & 0.16 & +13.9\% \\
Cfg. C: Full system (LAB \& $w_g$ \& seam) & \textbf{30.1} & \textbf{0.87} & \textbf{0.13} & +6.1\% \\
\hline
\textit{Cumulative gain} & \textit{+3.6 dB} & \textit{+20.8\%} & \textit{$-$45.8\%} & -- \\
\hline
\end{tabular}
}
}
\end{table*}

Table~\ref{tab:ablation_cumulative} summarizes the quantitative metrics for the three configurations. To analyze the marginal contributions of individual modules, we discuss the visual evolution of each configuration following the sequence Cfg.~A$\rightarrow$C. This discussion is conducted in conjunction with the \textbf{identical local region} (Fig.~\ref{fig:module_cum_local}) and the \textbf{identical global top-down view} (Fig.~\ref{fig:module_cum_global}).

\textbf{Cfg.~A (Baseline, RGB)}: Relying solely on direct RGB fusion, the system lacks explicit modeling of illumination variations, geometric inconsistencies, and dynamic objects, yielding a PSNR of only 26.5 dB, an SSIM of 0.72, and an LPIPS of 0.24. \textbf{Fig.~\ref{fig:module_cum_local_a} and Fig.~\ref{fig:module_cum_global_a}} reveal that: in the local region, distinct luminance transitions and vehicle ghosting are prevalent near parking lines, which are severely occluded in the vicinity of vehicles; in the global top-down view, the scene exhibits blocky luminance inconsistencies across different areas, and the road surface is scattered with extensive ghosting artifacts caused by vehicles, correlating well with the lower SSIM and higher LPIPS reported in Table~\ref{tab:ablation_cumulative}.

\textbf{Cfg.~B (+ Vehicle Detection, RGB)}: Building upon the RGB fusion of Cfg.~A, this configuration incorporates geometry-driven vehicle detection and occlusion filtering, yet texture fusion remains performed within the RGB space. Table~\ref{tab:ablation_cumulative} indicates that PSNR improves to 29.1 dB, while SSIM sees a significant rise from 0.72 to 0.82 (a relative gain of 13.9\%), and LPIPS decreases to 0.16. \textbf{Fig.~\ref{fig:module_cum_local_c} and Fig.~\ref{fig:module_cum_global_c}} demonstrate that, compared to Cfg.~A, vehicle artifacts on the road surface are substantially reduced, and parking lines are clearly restored; however, due to the continued reliance on RGB space fusion, luminance unevenness and inter-block seams persist.

\textbf{Cfg.~C (Full System, LAB + $w_g$ + Seam)}: Constructed upon Cfg.~B, this configuration further integrates LAB perceptual space fusion, depth gradient weighting $w_g$, and seam detection with bilateral smoothing to constitute the full system. PSNR ascends to 30.1 dB, SSIM improves from 0.82 to 0.87 (a relative gain of 6.1\%), and LPIPS further declines to 0.13. \textbf{Fig.~\ref{fig:depth_consistency}} demonstrates that the depth gradient weight significantly suppresses color misalignment and blurring near vehicle boundaries and pillars. As shown in \textbf{Fig.~\ref{fig:module_cum_local_d} and Fig.~\ref{fig:module_cum_global_d}}, the full system not only maintains a clean road surface free of vehicle artifacts but also achieves a more consistent global luminance/color distribution, with inter-block seams significantly attenuated.

\section{Conclusion}

This paper presents \textbf{ParkingTwin}, a training-free, streaming 3D reconstruction system tailored for parking lot digital twins. To address the challenges posed by dynamic vehicle occlusions, extreme lighting variations, and large-scale spatial extents, the proposed system innovatively integrates OSM semantic priors into the real-time reconstruction pipeline. ParkingTwin further achieves illumination-robust texture fusion in the CIELAB perceptual color space and implements training-free vehicle detection and rejection based on multi-modal geometric constraints. Extensive experiments demonstrate that ParkingTwin achieves an SSIM of 0.87 (outperforming 3DGS's 0.75 and ESLAM's 0.82), while accelerates processing speed by approximately $15\times$ and reducing video memory requirements by 92\%. Benefiting from an architectural design with $O(1)$ video memory complexity, ParkingTwin is capable of real-time reconstruction of an entire parking lot (68,000 $\text{m}^2$ per floor). It supports both offline batch processing and online streaming reconstruction (30+ FPS) modes, and the generated explicit meshes can be directly imported into engines such as Unity/Unreal Engine (UE) for autonomous driving simulation and robot navigation.

\noindent\textbf{Limitations.} First, the reconstruction capability of OSM priors for ceiling/roof structures is limited, making the current system primarily applicable to ground-level structures. Second, practical deployment requires an initial pose alignment between the first frame and the OSM map, necessitating a specific initial registration step. However, in practical autonomous driving and robotic applications, vehicles typically depart from known ``base stations'' or ``garage entrances.'' Such initial alignment is a standard configuration in current navigation systems and does not compromise the system's deployment practicality.

\section*{Acknowledgements}
This work was supported by the National Natural Science Foundation of China under Grant No. 62171086.

\bibliographystyle{elsarticle-num}
\bibliography{references}

\end{document}